\documentclass[lettersize,journal]{IEEEtran}
\usepackage{amsmath,amsfonts}
\usepackage{algorithm}
\usepackage{array}
\usepackage[caption=false,font=normalsize,labelfont=sf,textfont=sf]{subfig}
\usepackage{textcomp}
\usepackage{stfloats}
\usepackage{url}
\usepackage{verbatim}
\usepackage{graphicx}
\usepackage{cite}
\usepackage{times}
\usepackage{soul}
\usepackage[hidelinks]{hyperref}
\usepackage[utf8]{inputenc}
\usepackage{amsthm}
\usepackage{booktabs}
\usepackage{bbm}
\usepackage{amssymb}
\usepackage{pgfplots} 
\usepackage{times}
\usepackage{epsfig}
\usepackage{graphicx}
\usepackage{amsmath}
\usepackage{bbding}
\usepackage{pifont}
\usepackage{wasysym}
\usepackage{utfsym}
\usepackage{fontawesome}
\usepackage{tablefootnote}
\usepackage{colortbl}
\usepackage{xcolor} 
\usepackage{array,booktabs,makecell,tablefootnote}
\usepackage[flushleft]{threeparttable}
\usepackage{algorithmic}
\definecolor{gbypink}{rgb}{0.99, 0.91, 0.95}
\definecolor{tabgrey}{HTML}{E7EAEF}
\definecolor{tabblue}{HTML}{6B99C3}
\definecolor{codegreen}{rgb}{0,0.6,0}
\definecolor{codegray}{rgb}{0.5,0.5,0.5}
\definecolor{codepurple}{rgb}{0.58,0,0.82}
\definecolor{backcolour}{rgb}{1.0,1.0,1.0}

\usepackage[capitalize]{cleveref}
\crefname{section}{Section}{Sections}
\Crefname{section}{Section}{Sections}
\Crefname{table}{Table}{Tables}
\Crefname{figure}{Figure}{Figures}
\crefname{table}{Tab.}{Tabs.}

\makeatletter
\newcommand{\removelatexerror}{\let\@latex@error\@gobble}
\makeatother

\begin{document}

\title{Pixel-Level Domain Adaptation: \\ A New Perspective for Enhancing Weakly Supervised Semantic Segmentation}

\author{Ye Du,\   Zehua Fu,\  and Qingjie Liu, \IEEEmembership{Member, IEEE}
        % <-this % stops a space% <-this % stops a space
\thanks{Ye Du and Qingjie Liu are with the State Key Laboratory of Virtual Reality Technology and Systems, School of Computer Science and Engineering, Beihang University, Beijing, 100191, China.}
\thanks{Zehua Fu and Qingjie Liu are with the Hangzhou Innovation Institute, Beihang University, Hangzhou 310051, China.}
\thanks{Qingjie Liu is the corresponding author. E-mail: qingjie.liu@buaa.edu.cn.}
\thanks{Email of others:\{duyee, zehua\_fu\}@buaa.edu.cn}
}

\markboth{Journal of \LaTeX\ Class Files,~Vol.~14, No.~8, August~2021}%
{Shell \MakeLowercase{\textit{et al.}}: A Sample Article Using IEEEtran.cls for IEEE Journals}

\IEEEpubid{0000--0000/00\$00.00~\copyright~2021 IEEE}

\maketitle

\begin{abstract}
Recent attention has been devoted to the pursuit of learning semantic segmentation models exclusively from image tags, a paradigm known as image-level Weakly Supervised Semantic Segmentation (WSSS). 
Existing attempts adopt the Class Activation Maps (CAMs) as priors to mine object regions yet observe the imbalanced activation issue, where only the most discriminative object parts are located.
In this paper, we argue that the distribution discrepancy between the discriminative and the non-discriminative parts of objects prevents the model from producing complete and precise pseudo masks as ground truths.
For this purpose, we propose a Pixel-Level Domain Adaptation (PLDA) method to encourage the model in learning pixel-wise domain-invariant features.
Specifically, a multi-head domain classifier trained adversarially with the feature extraction is introduced to promote the emergence of pixel features that are invariant with respect to the shift between the source (\textit{i.e.}, the discriminative object parts) and the target (\textit{i.e.}, the non-discriminative object parts) domains.
In addition, we come up with a Confident Pseudo-Supervision strategy to guarantee the discriminative ability of each pixel for the segmentation task, which serves as a complement to the intra-image domain adversarial training.
Our method is conceptually simple, intuitive and can be easily integrated into existing WSSS methods.
Taking several strong baseline models as instances, we experimentally demonstrate the effectiveness of our approach under a wide range of settings.

\end{abstract}

\begin{IEEEkeywords}
Semantic segmentation, weakly supervised learning, domain adaptation, pseudo-labeling.
\end{IEEEkeywords}

\section{Introduction}

\IEEEPARstart{T}{he} success of semantic segmentation~\cite{long2015fully, zhang2014probabilistic, liang2015semantic, wei2016stc, zhao2017pyramid, fu2019stacked, wu2020cgnet, ding2020semantic, zhou2021gmnet, yang2021context, zhou2022rethinking, li2022attention, zhou2022weakly, wu2023conditional} heavily relies on large-scale annotated datasets.
However, manually annotating pixel-level labels is expensive and time-consuming, making the supervised approaches impractical in real-world contexts.
To surmount this constraint, numerous attempts~\cite{zhou2016learning, wei2017object, redondo2019learning, wang2020self, du2022weakly, chen2022self, zhou2022regional, lee2022threshold} have been directed toward the weakly supervised learning of semantic segmentation, where only image-level class labels are available during training.
Under this setting, generating precise semantic segmentation masks from image labels as pseudo ground truths becomes critical.

\begin{figure}[t]
    \centering
    \vspace{3pt}
\includegraphics[width=0.43\textwidth]{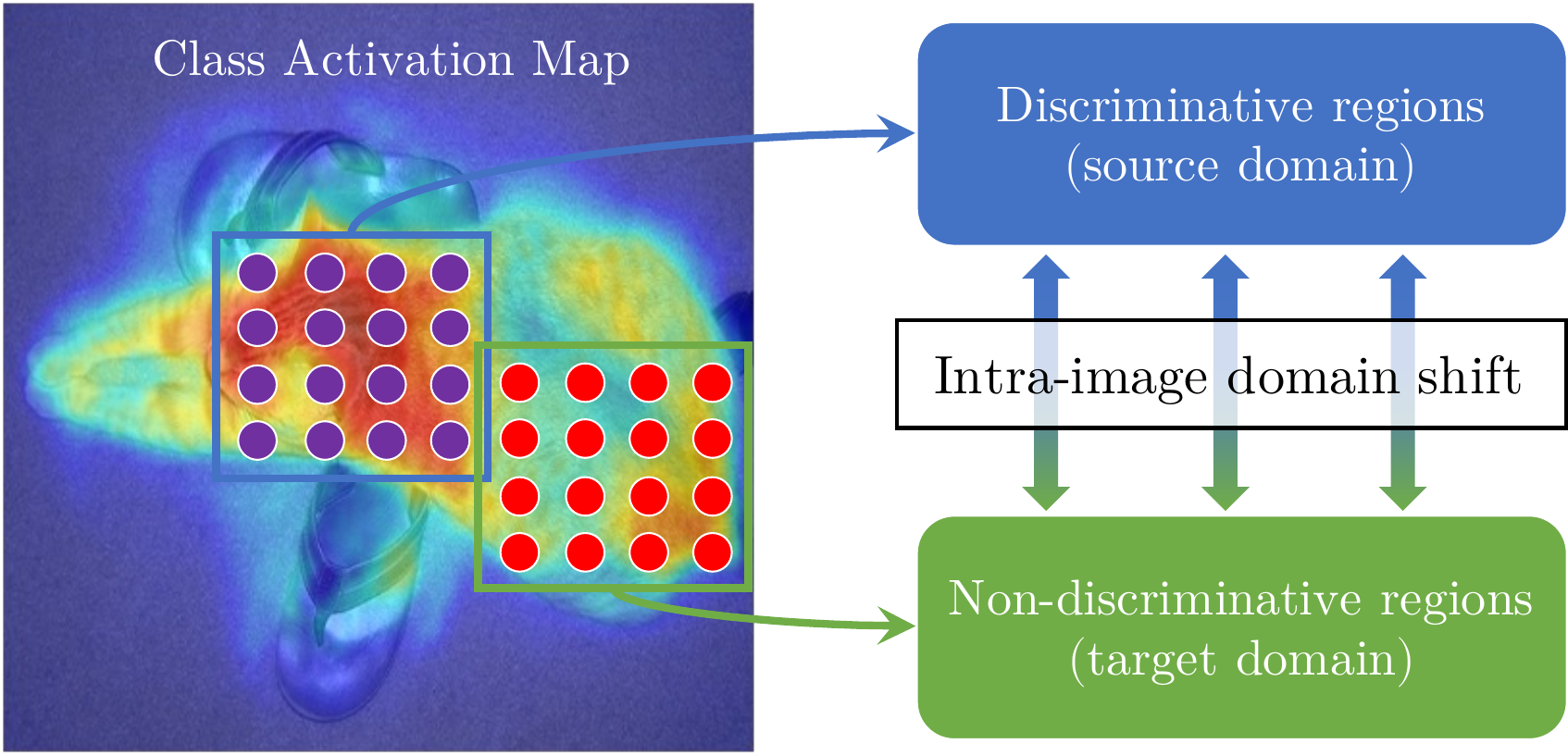}
    \caption{
        \textbf{Illustration of the motivation.} 
         The presence of a distribution discrepancy between the discriminative elements (\textit{e.g.}, the cat's head) and the non-discriminative components (\textit{e.g.}, the cat's body) of objects leads to a distinct activation pattern within the class activation map.
        To counteract this challenge, this paper proposes to explicitly align the pixel features of both types of regions in the image.
        \label{fig:motivation}
    }
\end{figure}

To obtain the desired segmentation maps, state-of-the-art works\cite{hou2018self, zhang2020causal, wang2020self, zhang2018adversarialCAMfirst, zhang2020reliability, zhang2021complementary, su2021context, zhou2022regional, jiang2022l2g, du2022weakly, chen2022self, zhou2021group} employ Class Activation Map (CAM)~\cite{zhou2016learning} to roughly localize the objects in an image.
In these methods, image-level labels are utilized to train a classifier, which in turn activates specific image regions for classification, thus yielding initial segmentation seeds. 
Such a mechanism bridges classification and segmentation, dominating existing weakly supervised semantic segmentation methods.
However, the CAM technique has long suffered from the issue of imbalanced activation~\cite{wei2017object, hou2018self, ahn2018learning, wang2020self, du2022weakly, chen2022self, jiang2022l2g, zhou2021group, chen2023multi}, where only the most discriminative parts of objects are highly activated and discovered.
This phenomenon arises as well-trained classifiers tend to internalize generalized and robust category patterns. 
Consequently, the quality of pseudo labels, along with the performance of the subsequent semantic segmentation, is unexpectedly limited.

\begin{figure}[t]
    \centering
\includegraphics[width=0.5\textwidth]
{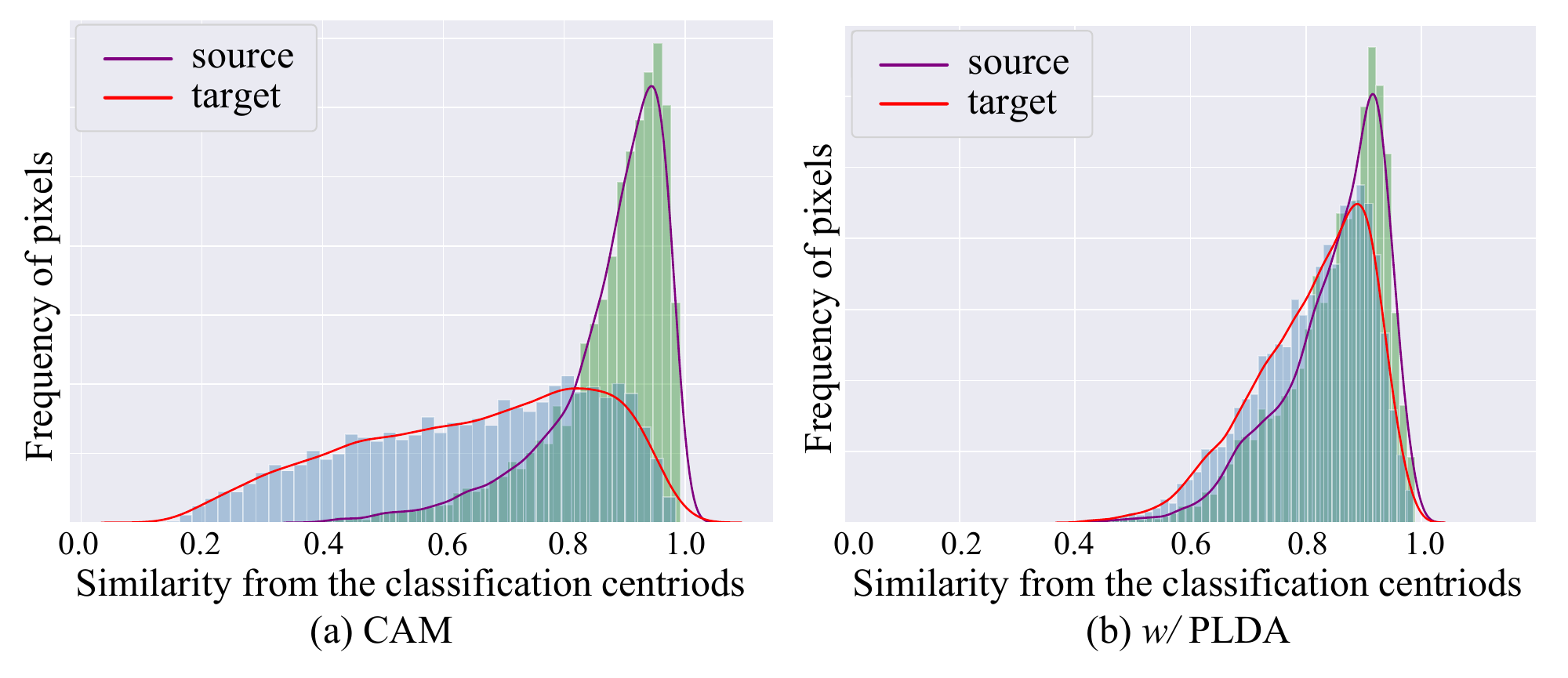}
    \caption{
    {
        \textbf{Illustration} of the distribution discrepancy between the most discriminative and less-discriminative regions.}
        The distribution is depicted by plotting the similarity scores between each pixel and its corresponding class-centroid, which is obtained by averaging all pixel features within the class.
        We denote the most discriminative regions as the ``source" and the others as the ``target".
        We use the PASCAL VOC 2012 \cite{everingham2010pascal} validation set that consists of $1,449$ images for plotting.
        To ensure an equal number of pixels in both domains, we sample $256$ pixels per class. 
        \label{fig:similarity}
    }
    \vspace{-1em}
\end{figure}

\IEEEpubidadjcol
In this work, we undertake a reexamination of the central crux of the aforementioned issue, approaching it through the lens of intra-image distribution discrepancy. 
By considering the example of a cat in \cref{fig:motivation}, the head region manifests distinct and informative cat-like attributes, whereas the body portion holds diminished import for object recognition. 
{
This clearly delineates a kind of pixel-level domain shift, where the features extracted from discriminative regions significantly differ from those of non-discriminative regions.
To provide further clarity, \cref{fig:similarity}a elucidates the distribution of normalized similarity scores between each pixel and its corresponding class-centroid, which indicates a tangible representation of the decision center learned by the CAM model within the feature space.
Upon closer examination, a pronounced distribution shift becomes evident, with the most discriminative pixels (depicted by the green bars) exhibiting markedly higher similarity scores compared to those non-discriminative counterparts (represented by the blue bars).
Consequently, we contend} that if we \textit{explicitly align the pixel features from the two distributions}, the CAM could cover more non-discriminative regions and produce more complete and precise pseudo masks to improve semantic segmentation.

Motivated by this observation, we propose a \textit{Pixel-Level Domain Adaptation} (PLDA) method, which aims at mitigating the discrepancy between the \textit{source} (discriminative regions) and the \textit{target} (remaining regions) domains. 
Intuitively, we draw inspiration from the principles of adversarial domain adaptation~\cite{ben2006analysis, ganin2015unsupervised, pei2018multi, wang2019weakly, hoffman2018cycada}, employing a domain discriminator operating within the pixel-level feature space to distinguish whether the pixel representation belongs to either the source or the target domain. 
Concurrently, the feature extractor undergoes adversarial training to confound the domain discriminator, thereby promoting the emergence of pixel features that are \textit{invariant} with respect to the two domains in the same image.
Considering the negative transfer problem~\cite{pei2018multi, li2018domain, deng2019cluster, wang2020classes} in domain adaptation, we then instantiate the domain discriminator with a multi-head classifier, where each head handles a specific category to avoid misalignment across classes.
In this way, the proposed PLDA manages to alleviate the imbalanced activation issue in the presence of intra-image domain shift regarding each category.

Furthermore, while domain adversarial training champions domain \textit{invariance}, the \textit{discriminability} of pixel features is equally vital to forestall degradation and facilitate segmentation. 
To this end, we further introduce a \textit{Confident Pseudo-Supervision} (CPS) strategy, which sieves reliable pseudo labels for guiding pixels affiliated with the source or target domain. Such a strategy enables a small expected error on both types of regions, thereby enhancing the semantic discriminability of individual pixel representations. 
Despite its simplicity, CPS synergistically complements the process of invariant feature learning, effectively improving the performance of the model.

{To summarize, our contribution is threefold:}

\begin{itemize}
    \item {
    We revisit the imbalanced activation issue in WSSS by examining the distribution discrepancy between discriminative and non-discriminative object regions. Analyzing the distribution of pixel features (as shown in \cref{fig:similarity}a) reveals a clear pixel-level domain shift, where features from discriminative regions significantly differ from those of non-discriminative regions. 
    We argue that explicitly aligning these pixel features can make CAM more complete and thus improve segmentation.
    }

    \item {To this end, we put forward a novel \textit{Pixel-Level Domain Adaptation} method to diminish the intra-image distribution discrepancy by prompting the pixel features to be both \textbf{(i)} \textit{domain invariant} against the intra-image shift (as shown in \cref{fig:similarity}b) through disentangled intra-image domain adversarial learning and \textbf{(ii)} \textit{semantic discriminative} for the segmentation task via a confident pseudo supervision strategy.}

    \item {We evaluate the proposed PLDA method on several strong baselines under a wide range of experimental settings (\textit{e.g.}, datasets and baselines). 
    The experimental results show that our method improves baseline models by large margins, demonstrating the effectiveness and generalizability.}
    
\end{itemize}

\section{Related Work}

\subsection{Weakly Supervised Semantic Segmentation}

Considerable strides have been achieved in the realm of Weakly Supervised Semantic Segmentation (WSSS) using solely image-level labels. The prevailing methods in this field generally adopt a multi-stage framework. Initially, these methods extract preliminary object seed areas from the Class Activation Map (CAM)\cite{zhou2016learning}, a byproduct of a classification model. Subsequently, these initial seeds undergo refinement to yield pseudo masks. Finally, these pseudo masks are harnessed for the fully supervised training of a semantic segmentation model, such as DeepLab\cite{liang2015semantic}.

Despite the progress, a noteworthy challenge surfaces in the application of CAM: it predominantly encompasses the most discriminative object regions, overlooking the broader object context. As a strategic remedy, adversarial erasing (AE) based techniques~\cite{wei2017object, hou2018self, zhang2018adversarialCAMfirst, kweon2021unlocking, yoon2022adversarial} have been introduced to compel models to attend to non-discriminative object areas. 
These approaches involve selectively erasing discriminative object parts, subsequently reintroducing the modified image into the network to uncover complementary regions.

{An alternative avenue of exploration~\cite{wang2020self, fan2020cianCIAN, wang2022looking, SPML, jiang2021online, du2022weakly, zhou2022regional, chen2022self, chen2023multi, zhang2023weakly, Hunting2024, chen2024spatial} centers on harnessing consistency priors, constraints and self-supervision signals to reinforce classification training.}
{Among these studies, exploring semantic consistency and supervision strategies across images shows promising performance \cite{fan2020cianCIAN, zhou2021group, wang2022looking, MCIC9873854}.}
{
To mitigate the co-occurrence issue,
Kweon et al. \cite{kweon2023weakly} propose to learn adversarially between the CAM network and a reconstructor, while Yang et al. \cite{yang2024separate} introduce a separate and conquer scheme targeting both the image and feature spaces.
}
{Moreover, rectification-based methods \cite{chen2023fpr, cheng2023out} aim to rectify CAM activations to facilitate more accurate object segmentation.
Some other studies \cite{chen2020weakly, chen2021end, wu2023conditional, Rong_2023_CVPR} explore boundary information to improve segmentation, producing more precise object boundaries.
}

{Moreover, augmenting supervision often entails additional signals such as saliency maps~\cite{zeng2019jointSaliency3, wu2021embedded, kim2021discriminative, lee2021railroad, xu2021leveraging, zhou2022regional, jiang2022l2g, EPSplus10120949}}, out-of-distribution images~\cite{lee2022weakly}, or {linguistic cues~\cite{xie2022cross, Lin_2023_CVPR, Yang_2024_WACV}.
These approaches greatly enhance the comprehensiveness of CAM and the performance of segmentation. }
Diverging from these methods, certain orthogonal techniques~\cite{ahn2018learning, ahn2019weakly, li2022towards, lee2022threshold, xie2022c2am, chen2022class} initially train a CAM model, leveraging it subsequently to refine or post-process via a distinct network.
Furthermore, a cluster of studies~\cite{li2022uncertainty, liu2022adaptive} has gravitated towards the enhancement of semantic segmentation training post the acquisition of pseudo-labels. 
{Recent explorations~\cite{ru2022learning, rossetti2022max, xu2022multi, ru2023token, AAAI24_Vit} also explore the potency of powerful Transformer networks~\cite{dosovitskiy2020image, touvron2021training} for WSSS, yielding encouraging outcomes.}

However, in distinction to prevailing pursuits, we propose to explicitly align the pixel representation between the image's discriminative and non-discriminative regions, presenting a new perspective beyond existing studies.

\begin{figure*}[t]
    \centering
    \includegraphics[width=1.0\textwidth]{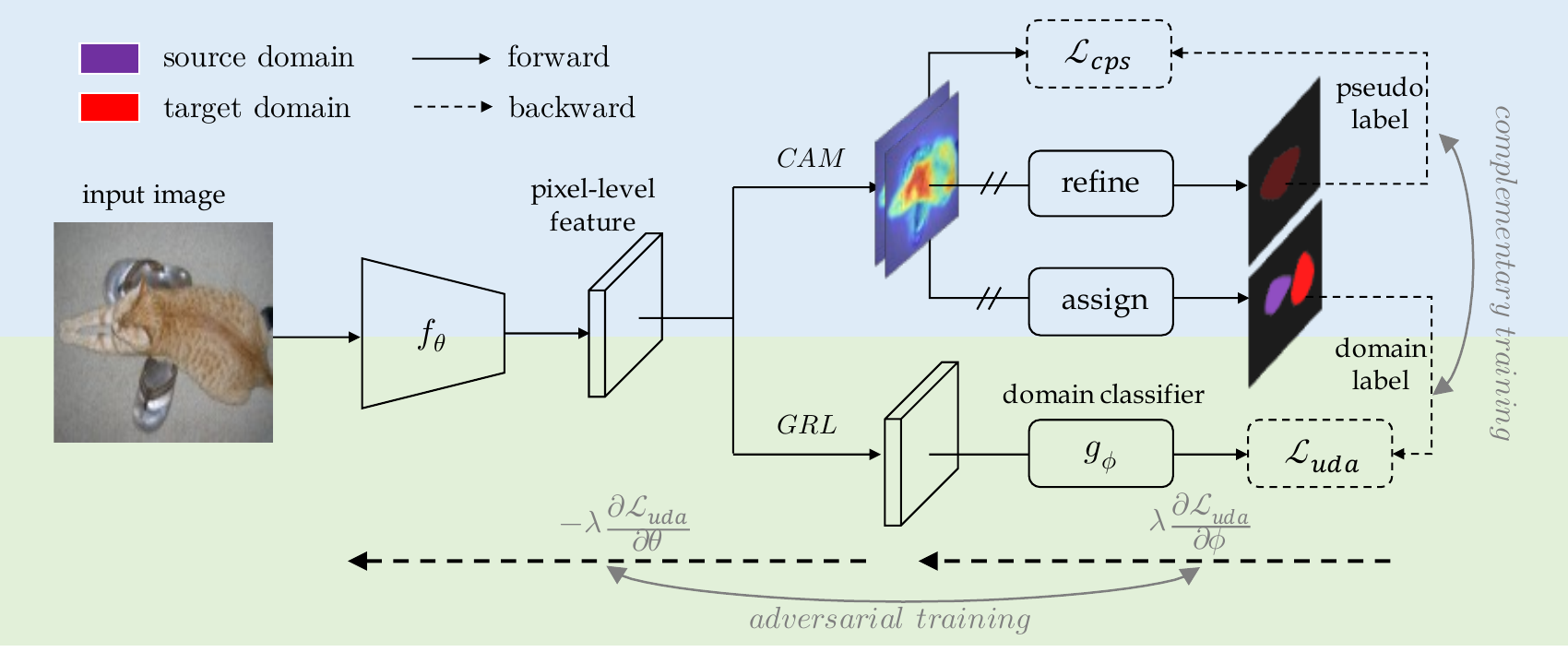}

    \caption{
        \textbf{Illustration of the PLDA framework.}
        Our method utilizes a domain classifier $g_{\phi}$ that trained adversarially with the feature extractor $f_{\theta}$ to learn \textit{domain invariant} features between the source and target domains.
        The source and target domain pixels are assigned dynamically according to CAM values.
        Additionally, we employ a confident pseudo-supervision mechanism on pixels associated with the source (or target) domain to ensure the essential discriminability for successful segmentation tasks, which operates in harmony with the domain adversarial training.
        ``GRL" means gradient reversal layer.
        }
    \label{fig:framework}
    \vspace{-1.3em}
\end{figure*}

\subsection{Unsupervised Domain Adaption}

In the absence of labeled data for a target domain, Unsupervised Domain Adaptation (UDA) provides an attractive solution given that labeled data of similar nature (\textit{i.e.}, the source domain) but from a different distribution are available.
A pivotal insight in UDA pertains to the minimization of distribution disparity across domains~\cite{wilson2020survey, geng2011daml}. 
Building upon this paradigm, domain adversarial training is introduced by Ganin \textit{et al}. \cite{ganin2015unsupervised} for visual recognition, entailing an adversarial objective to mitigate cross-domain incongruities. 
Diverse methods have sought to ameliorate this approach \cite{chen2019progressive, chen2019transferability, xu2020adversarial, pei2018multi, lin2020multi, hu2021adversarial}.
{For instance, progressive domain alignment \cite{chen2019progressive} is proposed to align the discriminative features across domains progressively and effectively.}
{Recent studies also extend the idea of domain alignment to encompass tasks spanning semantic segmentation~\cite{zou2018uda_self_training, vu2019advent, wang2019weakly, wang2020classes, cvpr2023UDA_seg} and object detection~\cite{munir2021ssal, jiang2021decoupled, aaai23_uda_det, pami_survey_det}.}

Notably, Zhu \textit{et al}. \cite{zhu2022weakly} have recently posited weakly supervised object localization as a distinct manifestation of domain adaptation. 
They assume a domain discrepancy between two feature spaces, \textit{i.e.}, the pixel-level features from the backbone and the image-level features after Global Average Pooling (GAP)~\cite{lin2013network}).
Then they utilize an MMD metric~\cite{dziugaite2015training} to minimize the distance, which can be viewed as narrowing the task gap between dense prediction and classification.
Yan \textit{et al}.~\cite{yan2021pixel} propose an intra-domain method for unsupervised domain adaptive semantic segmentation.
They utilize a continuous indexing technique in adversarial training to adapt the model from hard pixels in simple images to easy pixels in hard images. 

Our work is conceptually different from these works. 
Instead of focusing on feature space or intra-domain differences, our work minimizes the pixel-level discrepancy between \textit{discriminative} and \textit{non-discriminative} object regions to mitigate imbalanced activation for WSSS.

\subsection{Pseudo-labeling}

Another remarkable solution for learning from unlabeled data is the concept of pseudo-labeling {\cite{zou2018uda_self_training, song2020learning, zhang2021self, niu2022spice, xu2023cyclic, Padclip, energy}}, also known as self-training. 
The key insight of this approach involves generating pseudo-labels for unlabeled samples, thereby enabling the model to instruct itself. 
The efficacy of pseudo-labeling has rendered it a widely adopted technique in both semi-supervised learning~\cite{sohn2020fixmatch, berthelot2019mixmatch, tarvainen2017mean, chen2021semi, ke2022three} and domain adaptation~\cite{liu2021cycle, zheng2021rectifying, hoyer2022daformer, dulearning} fields.

In the area of WSSS, {some studies~\cite{zhang2020reliability, araslanov2020single,chen2021end, zhang2022end, ru2022learning, ru2023token, chen2023multi, yang2024separate, AAAI24_Vit}} develop end-to-end approaches that simultaneously perform classification and segmentation training.
Notably, these methods undertake online refinement of initial seed regions, customizing their strategies to ensure effective segmentation supervision. 
Illustrative techniques include dense Conditional Random Fields (CRF)\cite{zhang2020reliability}, Pixel Adaptive Mask Refinement (PAMR)\cite{araslanov2020single}, and Affinity from Attention (AFA)~\cite{ru2022learning}. Inspired by these works, this paper introduces a CPS strategy to ensure the semantic discriminative ability of pixel features, which complements the domain invariant learning.

\section{Methodology}

This section is organized as follows.
We commence this section by offering a succinct introduction to Class Activation Maps, thereby elucidating its intrinsic limitations. 
Subsequent to this introduction, we expound upon our pioneering Pixel-Level Domain Adaptation method, meticulously tailored to the domain of weakly supervised semantic segmentation. 
Building upon this foundation, we delve into the nuanced intricacies of the confidence pseudo-supervision strategy, revealing its pivotal role in our approach. To culminate, we encapsulate the comprehensive training objective and present the pseudo code that encapsulates the innovative methodology underpinning PLDA.

\subsection{Background}
\label{sec:method:background}

Image-level weakly supervised semantic segmentation  aims to acquire a semantic segmentation model using a dataset denoted as $\mathcal{D}=\left\{ (\mathbf{x}_i, \mathbf{y}_i)\right\}_{i=1}^{N}$. In this dataset, $\mathbf{x}_i \in \mathbb{R}^{H \times W \times 3}$ represents an RGB image, while $\mathbf{y}_i \in \{0,1\}^C $ signifies the multi-hot label indicating the presence of objects within $\mathbf{x}_i$. The values of $N$ and $C$ correspond to the sample count and class count, respectively.
This task proves to be formidable due to the inherent inability of image-level labels to convey object spatial localization information, which is essential for accurate semantic segmentation.

Thankfully, the Class Activation Map offers a solution to bridge this gap. The CAM identifies specific image regions pivotal for discerning a given category. Concretely, for an image $\mathbf{x}$ with its image-level label $\mathbf{y}$, a Convolutional Neural Network $f$ with parameters $\theta$ is employed to extract the feature map $\mathbf{z} \in \mathbb{R}^{ h \times w \times D}$, where $D$ represents the feature dimension, and $h\times w$ signifies the spatial dimensions.
Subsequently, the class activation map for category $c$ can be derived as follows:
\begin{equation}
\begin{aligned}
\mathbf{m}_c &= \text{ReLU}\left(\mathbf{w}_c \mathbf{z} ^\top \right),
\end{aligned}
\end{equation}

where $\mathbf{w} \in \mathbb{R}^{C\times D} $ denotes the weight of a classification head integrated atop $f$.

Afterwards, the class activation maps are typically refined to generate the pseudo mask, which serves as the ground truth to train a semantic segmentation model.

\vspace{3pt}
\textbf{Limitation of CAM}. 
Despite the effectiveness, CAM confronts the predicament of imbalanced activation. 
The crux of this issue lies in the distribution discrepancy between discriminative and non-discriminative object regions, which impedes the generation of uniform activation values for all pixels within the object.
While considerable progress has been made to improve the integrity of CAM, few studies have addressed the imbalanced activation from this perspective.

\subsection{Pixel-Level Domain Adaptation for WSSS}
\label{sec:method:plda_framework}

As previously discussed, our objective is to explicitly align the pixel representations between discriminative and non-discriminative object regions to encourage similar activation in CAM.
To this end, we formulate WSSS as an intra-image domain adaptation challenge. 
In this context, the most discriminative object regions, such as the head of a cat, are treated as the \textit{source} domain, while the less discriminative object regions, such as the body of a cat, constitute the \textit{target} domain. 
Our proposed approach, denoted as \textit{Pixel-Level Domain Adaptation}, is designed to achieve the required alignment. 
\cref{fig:framework} illustrates the PLDA framework.
Next, we elaborate on the details of our approach.

\textbf{Intra-image Domain Adversarial Training}:
As depicted in \cref{fig:framework}, our PLDA framework comprises a feature extractor $f_\theta$ and a domain classifier $g_\phi$, which follows the idea of domain adversarial training~\cite{ganin2015unsupervised, ganin2016domain}. Notably, our approach operates at the pixel level \textit{within} an image.

Concretely, denoted by $\mathcal{D}_s=\{\mathbf{x}_i^s \}_{i=1}^{N_s}$ the pixels of the source domain, and correspondingly by $\mathcal{D}_t=\{\mathbf{x}_j^t \}_{j=1}^{N_t}$ the one of the target domain, the domain classifier is trained to distinguish $\mathcal{D}_s$ from $\mathcal{D}_t$.
In contrast, the feature extractor is optimized 
simultaneously to confuse the domain classifier, which urges learning domain invariant pixel features.
Formally, while the parameters $\phi$ of the domain classifier are learned by minimizing a loss of domain classification $\mathcal{L}_{uda}$, the parameters $\theta$ of the feature extractor are optimized to maximize it.
We define $\mathcal{L}_{uda}$ as a binary cross-entropy loss:

\begin{equation}
\begin{aligned}
    \mathcal{L}_{uda}(\theta, \phi) & = \frac{1}{|\mathcal{D}_s| + |\mathcal{D}_t|}
    \sum_{\mathbf{x}_i \in (\mathcal{D}_s \cup \mathcal{D}_t) } \ell_{ce}\left(g_\phi \circ f_\theta(\mathbf{x}_i), \mathbf{d}_i \right),
\end{aligned}
\label{eq:uda_loss}
\end{equation}

where $\mathbf{d}_i$ is the one-hot domain label indicating the domain of the pixel $\mathbf{x}_i$. 
$\ell_{ce}$ is the standard cross-entropy loss, \textit{i.e.}, $\ell_{ce}(\mathbf{p}, \mathbf{y}) = - \mathbf{y}^\top\log\mathbf{p}$.
Then, the adversarial training can be viewed as playing a two-player minimax game with respect to $\mathcal{L}_{uda}$, that is
% %\vspace{-5pt}
\begin{equation}
%\vspace{-5pt}
\begin{aligned}
    \min_{\phi} \max_{\theta} \mathcal{L}_{uda}(\theta, \phi).
\end{aligned}
%\vspace{-3pt}
\end{equation}
% %\vspace{-10pt}

After training convergence, the parameters deliver a saddle point, which can be written as
% %\vspace{-5pt}
\begin{equation}
%\vspace{-5pt}
\begin{aligned}
    \hat{\phi} &= \arg \min_{\phi} \mathcal{L}_{uda} (\hat{\theta}, \phi),\\
    \hat{\theta} &= \arg \max_{\theta} \mathcal{L}_{uda} (\theta, \hat{\phi}).
\end{aligned}
%\vspace{-2pt}
\end{equation}
% %\vspace{-10pt}

Note that we omit the CAM training loss here for simplicity.
In addition, we follow~\cite{ganin2015unsupervised} to use a Gradient Reversal Layer (GRL) to implement adversarial training.
As shown in \cref{fig:framework}, GRL acts as an identity function during forward propagation but flips the sign of the gradient during the backward pass.
Thereby, our approach can be trained end-to-end and easily integrated into existing weakly supervised semantic segmentation methods.

\begin{figure}[t]
    \centering
    \vspace{15pt}
    \includegraphics[width=0.43\textwidth]{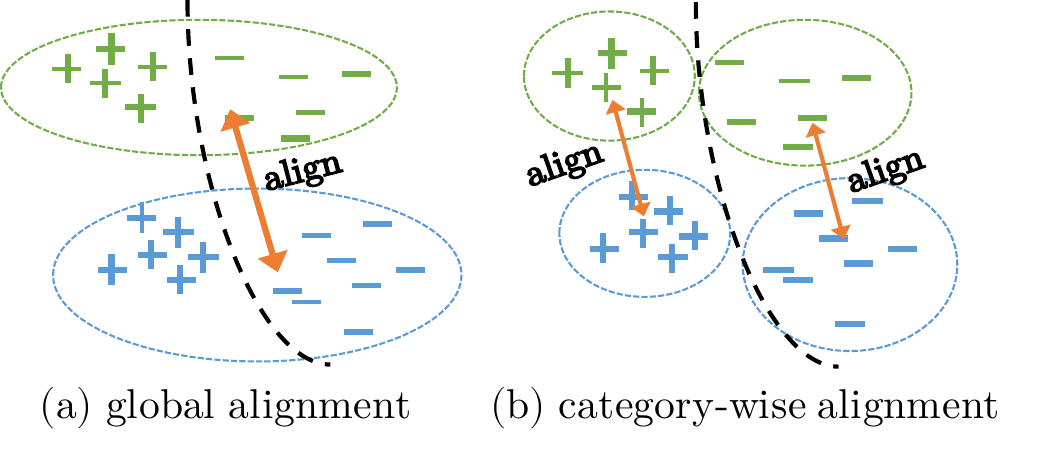}
    \caption{
        \textbf{Comparison of the two alignment ways.} 
        (a) Global method aligns the marginal distributions without considering the per-class information.
        (b) Category-wise method aligns the class-conditional distributions to reduce the mismatching between classes.
        ``$+$" and ``$-$" indicate two categories. Green and blue colors indicate samples of source and target domains, respectively.
        }
    \label{fig:alignment}
    %\vspace{-12pt}
\end{figure}

\begin{figure}[t]
    \centering
    \includegraphics[width=0.43\textwidth]{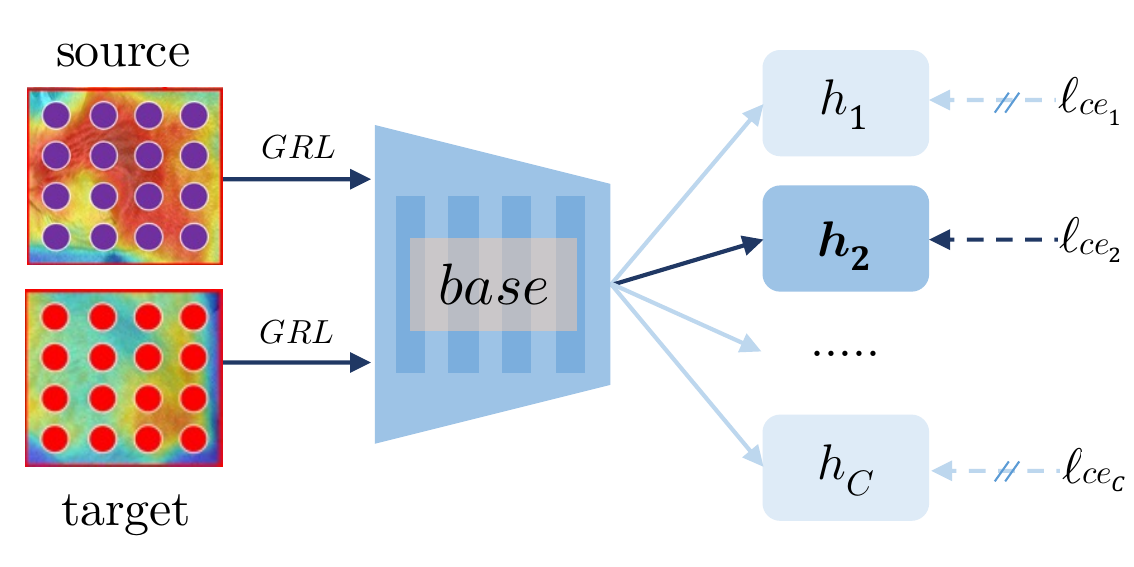}
    %\vspace{-9pt}
    \caption{
        \textbf{Illustration of the multi-head domain classifier.} 
        The $g_{\phi}$ is instantiated with a shared \textit{base} network and $C$ classifier heads, where each head (\textit{e.g.}, $h_2$ in this instance) distinguishes source and target pixels for a specific category.
        This simple design avoids the misalignment across classes.
        }
    \label{fig:multihead}
\end{figure}

\subsection{Domain Adaptation Disentanglement}

Intra-image domain adversarial training facilitates invariant pixel feature learning by aligning the marginal distributions globally between the two domains.
However, as shown in~\cref{fig:alignment}(a), the global alignment ignores class-conditional structures and thus bears the risk of misalignment between classes~\cite{wang2020classes}.
Moreover, this paradigm tends to disproportionately focus on high-frequency classes while neglecting low-frequency ones, such as categories with fewer samples or smaller object sizes~\cite{wang2019class, deng2019cluster}.

Thus, inspired by ~\cite{pei2018multi, hong2018conditional, long2018conditional}, we further introduce a multi-head domain classifier as the discriminator to disentangle the domain adaptation.
Specifically, as shown in~\cref{fig:multihead}, the multi-head domain classifier is composed of a shared base network and $C$ binary classification heads, where each head is responsible for the domain discrimination of a specific category.
During training, we utilize CAM to indicate the category of each pixel and assign it a classification head accordingly.
Then, a category-specific domain adaptation shown in \cref{fig:alignment}(b) can be achieved by a slight modification to \cref{eq:uda_loss}, that is,

\begin{equation}
\label{eq:multiuda}
\begin{aligned}
    \mathcal{L}_{uda}(\theta, \phi) = & \frac{1}{|\mathcal{D}_s| + |\mathcal{D}_t|}
    \sum_{\mathbf{x}_i \in (\mathcal{D}_s \cup \mathcal{D}_t) } \sum_{c} \\
    &\mathbbm{1}[\arg\max \mathbf{m}_i=c]\ell_{ce_c}\left(g_\phi \circ f_\theta(\mathbf{x}_i), \mathbf{d}_i \right),
\end{aligned}
\end{equation}
where $\mathbf{m}_i \in \mathbb{R}^{C}$ is the class activation distribution of pixel $\mathbf{x}_i$, $\mathbbm{1}[\cdot]$ is an indicator function.
In this way, we manage to disentangle the multi-class domain adaptation and reduce the misalignment.
Algorithm 1 summarizes the specific calculation process of the proposed PLDA method.

\begin{figure}[!t]
    \label{alg:PLDA}
    \removelatexerror
    \begin{algorithm}[H]
        \caption{Pixel-Level Domain Adaptation for WSSS}
        \begin{algorithmic}[1]
            \REQUIRE Epoch $E$; Dataset $D$; Model $f_{\theta}(\cdot)$, Domain classifier $g_{\phi}(\cdot)$; CAM generation process $CAM$; Threshold $\alpha$.

            \ENSURE Optimized model $f_{\theta^{*}} (\cdot)$. 
    
            \FOR{each $t \in [0,E]$}
        
            \STATE Sample a data batch $\mathbf{x} \sim D$;
            \STATE $\mathbf{z} = f_{\theta}(\mathbf{x})$; 
            \STATE $\mathbf{m} = CAM(\mathbf{z})$;
            \STATE $\widetilde{\mathbf{x}} = [\mathbf{1} - (\mathbf{m}>\alpha)]  \odot \mathbf{x}$;
            \STATE $\widetilde{\mathbf{z}} = f_{\theta}(\widetilde{\mathbf{x}})$;
            \STATE $\widetilde{\mathbf{m}} = CAM (\widetilde{\mathbf{z}})$;
            \STATE $\mathbf{d}_s, \mathbf{d}_t = \mathbf{m} > \alpha, \widetilde{\mathbf{m}} > \alpha $;
            \STATE $\mathbf{d} = \text{concat}(\mathbf{d}_s, \mathbf{d}_t)$;
            \STATE $\mathcal{D}_s, \mathcal{D}_t = \{\mathbf{x}_i | \mathbf{m}_i > \alpha \},\{\mathbf{x}_i | \widetilde{\mathbf{m}}_i > \alpha \}$;
            \STATE $\mathcal{L}_{uda}(\theta, \phi) = \frac{1}{|\mathcal{D}_s| + |\mathcal{D}_t|}\sum_{\mathbf{x}_i \in (\mathcal{D}_s \cup \mathcal{D}_t) } \sum_{c}$
            \STATE \ \ \ \ \ \ \ \ \ \ \ \ \ \ \ \ $\mathbbm{1}[\arg\max\mathbf{m}_i=c]\ell_{ce_c}\left(g_\phi(\mathbf{z}_i), \mathbf{d}_i \right)$;
            
            \STATE Optimize $\min_{\phi} \max_{\theta} \mathcal{L}_{uda}(\theta, \phi)$;

            \STATE Update parameters $\theta$ and $\phi$.
            \ENDFOR
        \STATE Set ${\theta^*} \leftarrow {\theta}$.
        \RETURN $f_{\theta^*}(\cdot)$
        \end{algorithmic}
    \end{algorithm}
    \vspace{-2em}
\end{figure}

\subsection{Domain Assignment}

Recall that our primary motivation is to explicitly align the pixel representation between the discriminative and non-discriminative object regions.
In the preceding parts, we present our proposed solution, namely PLDA.
However, a fundamental problem still demands resolution: the precise identification of pixels belonging to the respective source and target domains.

To establish the domain label corresponding to each individual pixel in \cref{eq:multiuda}, we introduce an innovative \textit{Masked Assignment} (MaskAssign) strategy.
Specifically, for a given image $\mathbf{x}$, we forward it into $f_\theta$ to obtain the CAM $\mathbf{m}$.
Similar to the aforementioned procedure, we first set a threshold $\alpha$ to filter the most discriminative regions as the source domain, \textit{i.e.},
$\mathcal{D}_s = \{\mathbf{x}_i | \mathbf{m}_i > \alpha \}_{i=1}^{H\times W}$.
Then, we erase the source domain pixels to obtain a masked image, that is,

\begin{equation}
    \widetilde{\mathbf{x}} = [\mathbf{1} - (\mathbf{m}>\alpha)]  \odot \mathbf{x}.
\end{equation}

Afterwards, $\widetilde{\mathbf{x}}$ is input again into $f_\theta$ to get a masked CAM $\widetilde{\mathbf{m}}$, which is utilized to determine the target pixels.
Concretely, we treat pixels whose masked CAM values are still larger than the threshold $\alpha$ as the target domain, \textit{i.e.}, $\mathcal{D}_t = \{\mathbf{x}_i | \widetilde{\mathbf{m}}_i > 
\alpha \}_{i=1}^{H\times W}$.
This strategy is inspired and evidenced by adversarial erasing works~\cite{wei2017object, hou2018self}, which prove that a classifier is capable of identifying subactivation regions by erasing the most discriminative parts of the object in the image.
As such, our method is expected to continuously align the pixel representations between the most discriminative and the other regions along the training.

\subsection{Confident Pseudo-Supervision}
\label{sec:method:cps}
Domain adversarial training has been criticized for potentially impairing the discriminative ability of features~\cite{chen2019transferability, zhang2021prototypical}.
To mitigate this concern, we further put forward \textit{Confident Pseudo-Supervision} (CPS), which leverages reliable pseudo-labels to supervise the source and target domain pixels.
Concretely, as illustrated in \cref{fig:framework}, we first refine the CAM map to get the pseudo-label distribution $\mathbf{p} \in \mathbb{R}^{h\times w\times C}$, then impose supervision on the \textit{source} domain pixels whose pseudo-label confidence larger than a threshold $\beta$, that is
\begin{equation}
    \label{sourcece}
    \begin{aligned}
            \mathcal{L}_{cps_s} =& \frac{1}{\sum_j\mathbbm{1}[\mathbf{x}_j \in \mathcal{D}_s; \max\mathbf{p}_j > \beta]} \sum_{i} \\
            &\mathbbm{1}[\mathbf{x}_i \in \mathcal{D}_s; \max\mathbf{p}_i > \beta] \ell_{ce}(f_{\theta}(\mathbf{x})_i, \arg\max\mathbf{p}_i),
    \end{aligned}
\end{equation}
where $i$ indexes each pixel and the maximum value of pseudo prediction is used as the confidence indicator.
Besides, we follow~\cite{araslanov2020single} to set $\beta$ as a dynamic threshold, namely
$\beta = \beta'\cdot \max\{\mathbf{p}_{j,c}\}_{j=1}^{h\times w}$, where $c=\arg\max\mathbf{p}_i$ and $\beta'$ $\in$ $(0,1)$ is a hyper-parameter.
Symmetrically, we refine the masked CAM $\widetilde{\mathbf{m}}$ to obtain the prediction $\widetilde{\mathbf{p}}$, which serves as pseudo truth to supervise the target domain pixels, that is
\begin{equation}
    \label{targetce}
    \begin{aligned}
            \mathcal{L}_{cps_t} = & \frac{1}{ \sum_j\mathbbm{1}[\mathbf{x}_j \in \mathcal{D}_t; \max\widetilde{\mathbf{p}}_j > \beta]} \sum_{i} \\ 
            & 
            \mathbbm{1}[\mathbf{x}_i \in \mathcal{D}_t; \max\widetilde{\mathbf{p}}_i > \beta] \ell_{ce}(f_{\theta}(\mathbf{x})_i, \arg\max\widetilde{\mathbf{p}}_i).
    \end{aligned}
\end{equation}

Finally, the CPS loss can be written as $\mathcal{L}_{cps} = \mathcal{L}_{cps_s} + \mathcal{L}_{cps_t}$.
It is worth noting that CPS facilitates small cross-entropy errors on both domains, such that the discriminative information of each pixel representation can be retained.

\subsection{Total Training Objective}
\label{sec:method:objective}
To summarize, the total loss of our approach is the combination of classification training loss, the intra-image domain adversarial training loss  and the confident pseudo-supervision loss, that is
\vspace{-0.5em}
\begin{equation}
    \begin{aligned}
            \mathcal{L}_{total} = \mathcal{L}_{cls} + \mathcal{L}_{uda} +  \mathcal{L}_{cps},
    \end{aligned}
\end{equation}
where $\mathcal{L}_{cls}$ is defined as a multi-label cross-entropy loss~\cite{wang2020self} on the aggregation of CAM.
Note that unlike general gradient descent updates, the domain classifier $g_\phi$ and the feature extractor $f_\theta$ are trained adversarially.

\begin{table}[t]
  
\centering
    \caption{\textbf{Evaluation} on class activation map. 
    The mIoU (\%) values on training and validation sets of PASCAL VOC 2012 are reported. 
    Our proposed PLDA brings consistent improvements on four baseline methods.
    }
    \label{tab:cam-voc}
    %\vspace{3pt}
\renewcommand{\arraystretch}{1.05}%{1.05}
% {
\scalebox{0.97}{
\begin{tabular}{l|l|l|c|c}
\toprule
Method & Publication &Backbone & Train & Validation  
\\ \midrule%[0.8pt]
CONTA~\cite{zhang2020causal} & NeurIPS'20    & ResNet38 & 56.2  & - \\
EDAM~\cite{wu2021embedded} & CVPR'21 & ResNet38 & 52.8 & 51.8 \\
CDA~\cite{su2021context}     & ICCV'21    & ResNet38      & 58.4  & - \\
RIB~\cite{lee2021reducing}     & NeurIPS'21 & ResNet50      & 56.5  & 55.4 \\
AdvCAM~\cite{lee2022anti}  & TPAMI'22  & ResNet50  & 55.6  & 54.6 \\
CLIMS~\cite{xie2022cross}  & CVPR'22 & ResNet50     & 56.6  & - \\
W-OoD~\cite{lee2022weakly} & CVPR'22 & ResNet38     & 55.9  & - \\
MCTformer~\cite{xu2022multi} & CVPR'22    & DeiT-S     & 61.7  & 60.0 \\
ESOL~\cite{li2022expansion}    & NeurIPS'22     & ResNet50  & 53.6  & - \\
AEFT~\cite{yoon2022adversarial} & ECCV'22 & ResNet38 &56.0 &- \\

{FPR\tiny{\textit{w}/SEAM}\scriptsize{~\cite{chen2023fpr}}} & {ICCV'23} & {ResNet38} & {57.0} & {-}  \\

{ACR~\cite{kweon2023weakly}} & {CVPR'23} & {ResNet38} & {60.3} & {-}  \\

{SSC~\cite{chen2024spatial}} & {TIP'24} & {ResNet50} & {58.3} & {-} \\

{BAS~\cite{zhai2024background}} & {IJCV'24} & {ResNet50} & {57.7} & {-} \\

\midrule
$\mbox{PSA}^{+}$~\cite{ahn2018learning}  & CVPR'18 & ResNet38 & 50.3  & 48.3    \\
\cellcolor{tabblue!20} \ \ \ \ +\ PLDA  &\cellcolor{tabblue!20}--       & \cellcolor{tabblue!20}ResNet38 & \cellcolor{tabblue!20}{53.9}  & \cellcolor{tabblue!20}{51.4}    \\ 
SEAM~\cite{wang2020self}   & CVPR'20 & ResNet38 & 55.4  & 52.5 \\
\cellcolor{tabblue!20} \ \ \ \ +\ PLDA   & \cellcolor{tabblue!20}--        & \cellcolor{tabblue!20}ResNet38 & \cellcolor{tabblue!20}{59.1}  & \cellcolor{tabblue!20}{56.3} \\
SIPE~\cite{chen2022self}     & CVPR'22 & ResNet50 & 58.7  & 57.5 \\
\cellcolor{tabblue!20}\ \ \ \ +\ PLDA     & \cellcolor{tabblue!20}--       & \cellcolor{tabblue!20}ResNet50 & \cellcolor{tabblue!20}{60.1}  & \cellcolor{tabblue!20}{59.6} \\ 
PPC~\cite{du2022weakly}     & CVPR'22 & ResNet38 & 61.5  & 58.4 \\
\cellcolor{tabblue!20}\ \ \ \ +\ PLDA    & \cellcolor{tabblue!20}--      & \cellcolor{tabblue!20}ResNet38 & \cellcolor{tabblue!20}\textbf{62.5}  & \cellcolor{tabblue!20}\textbf{60.1} \\
\bottomrule%[1pt]
\end{tabular}
}
\end{table}

\section{Experiment}

\subsection{Setup}

\subsubsection{Dataset}

Our method is subjected to a rigorous evaluation that includes the PASCAL VOC 2012 dataset~\cite{everingham2010pascal} and the MS COCO 2014 dataset~\cite{lin2014microsoft}.

The PASCAL VOC 2012 dataset consists of $21$ classes, including $20$ distinct objects in addition to the background. 
This dataset is amply endowed with $1,464$, $1,449$, and $1,456$ images allocated across the training, validation, and test sets, respectively. 
Following the common practice in semantic segmentation, we adopt from \cite{hariharan2011semantic} the augmented set that consists of $10,582$ images to train the WSSS models.

MS COCO 2014 has $81$ classes ($80$ objects and the background) with $80\text{k}$ and $40\text{k}$ images for training and validation, respectively.
Consistent with the established conventions of the WSSS methodology~\cite{lee2021railroad, chen2022self}, images without target classes are judiciously excluded from consideration.

\subsubsection{Evaluation Metric}
Following previous work, we report the Mean Intersection over Union (mIoU) value for evaluation.

\subsubsection{Implementation Details of CAM Training} 
We adopt four high-performance WSSS methods (PSA, SEAM, PPC and SIPE) as baselines.
Incorporating our PLDA approach into baseline models involves employing a straightforward MLP network to serve as the domain classifier. 
For this integration, we select the PAMR module~\cite{araslanov2020single} to refine the CAM to derive pseudo labels, although alternative enhancement methods, such as denseCRF~\cite{zhang2020reliability}, can also be considered. 
Our choice of PAMR is motivated by its training efficiency. 
During training, we adhere to the training settings of the baseline models to ensure equitable comparisons. The specifics are as follows.

\textit{Enhancing the SEAM Baseline:}
To extend the SEAM baseline, we adopt the ResNet38~\cite{wu2019wider}, pre-trained on ImageNet~\cite{deng2009imagenet}, as the backbone network with an output stride of 8. 
Training extends for 8 epochs, utilizing the SGD optimizer with a batch size of 8 and an initial learning rate of 0.01. 
For image inputs, we follow SEAM~\cite{wang2020self}, rescaling them within the range of $[448, 768]$ by their longest edge and cropping to $448 \times 448$. 
Additionally, data augmentation employs random horizontal flipping and color jitter. 
The PLDA model is trained on two NVIDIA GeForce RTX 2080Ti GPUs with a batch size of 8 for 8 epochs. 
A learning rate of 0.01 is employed initially, following which the poly policy $lr_{t} = 0.01 \times (1-{t}/{T})^{\gamma}$ is applied for learning rate decay, with $\gamma$ set to 0.9. 
When calculating $\mathcal{L}_{uda}$, we follow~\cite{araslanov2020single} to adopt a weighted multi-class cross-entropy loss.
We implement the domain classifier as a simple MLP network with $C$ linear classifier heads.
To be specific, the architecture of \textit{base} network is
$1\times 1\ \text{Conv2d} \rightarrow \text{ReLU} \rightarrow \text{Dropout} \rightarrow 1\times 1\ \text{Conv2d} \rightarrow \text{ReLU} \rightarrow \text{Dropout}$.
Besides, the classifier heads share the same architecture, \textit{i.e.}, a single $1\times 1$ convolution layer.
During inference, we adopt multi-scale and flipping as test time augmentation.
Following SEAM~\cite{wang2020self}, we traverse all background threshold options and give the best mIoU value for CAM evaluation.

\textit{Enhancing the SIPE Baseline:}
Expanding upon the SIPE baseline, we use the ImageNet-pretrained ResNet50~\cite{he2016deep} as the backbone network with an output stride of 16. Image inputs are randomly rescaled within $[320, 640]$ by their longest edge and cropped to $512 \times 512$. Training for the PLDA model involves using two NVIDIA GeForce RTX 2080Ti GPUs with a batch size of 16 for 5 epochs.

\textit{Extending the PPC Baseline:}
In the case of the PPC baseline, the setting is similar to SEAM, but source images are resized to $128 \times 128$ for target branch input. Training the PLDA model occurs on three NVIDIA GeForce RTX 2080Ti GPUs, with three images per GPU for 8 epochs.

\begin{table}[t]
  
\centering
\caption{\textbf{Evaluation} on class activation map. 
The mIoU (\%) on the training set of MS COCO 2014 is reported. 
Our result is marked in bold.
``*" means our re-implementation.
``I." and ``S." mean methods using image-level and saliency supervision, respectively.
}
\label{tab:cam-coco}
\vspace{-5pt}
\renewcommand{\arraystretch}{1.0}
\scalebox{0.9}{{\begin{tabular}{l|l|l|c|c}
\toprule%[1pt]
Method & Publication & Backbone & Supervision & mIoU \\ \midrule%[0.8pt]
EPS~\cite{lee2021railroad}    & CVPR'21     & ResNet38 & I.+S.  & 37.2 \\
ESOL~\cite{li2022expansion}    & NeurIPS'22     & ResNet50 & I.  & 35.7 \\
SIPE~\cite{chen2022self}   & CVPR'22     & ResNet50 & I.    & 35.0 \\
{BAS~\cite{zhai2024background}}& IJCV'24     & ResNet50 & I.    & 36.9 \\

\midrule
SIPE*~\cite{chen2022self}  & CVPR'22     & ResNet50 & I.    & 35.6 \\

\cellcolor{tabblue!20} \ \ \ +\ PLDA   & \cellcolor{tabblue!20}--           & \cellcolor{tabblue!20}ResNet50 & \cellcolor{tabblue!20}I.    & \cellcolor{tabblue!20}\textbf{37.3} \\
\bottomrule
\end{tabular}}}
\end{table}

\subsubsection{Implementation Details of Segmentation Training}

In our segmentation training phase, we capitalize on the generated pseudo-masks to serve as ground truths for training DeepLabV1 models on the PASCAL VOC 2012 dataset~\cite{liang2015semantic}. Specifically, input images are subject to random rescaling within the interval $[0.5, 1.5]$ relative to the original dimensions, followed by cropping to $448 \times 448$ dimensions. To further enhance dataset diversity, we introduce random horizontal flipping and color jitter as data augmentation strategies.
The DeepLabV1 models are trained on a single NVIDIA V100 GPU, utilizing a batch size of 10 for 20,000 iterations, which approximates 20 epochs. 
During the inference phase, we use multi-scale and flipping techniques as part of test time augmentation. 
Moreover, we leverage the conventional dense CRF to further refine the segmentation outputs.
Furthermore, we train DeepLabV1 models on the MS COCO 2014 dataset. In this case, the DeepLabV1 training regimen encompasses 80,000 iterations (about 10 epochs) on a single NVIDIA V100 GPU with a batch size of 10.

\begin{figure}[t]
    \centering
    \includegraphics[width=0.48\textwidth]{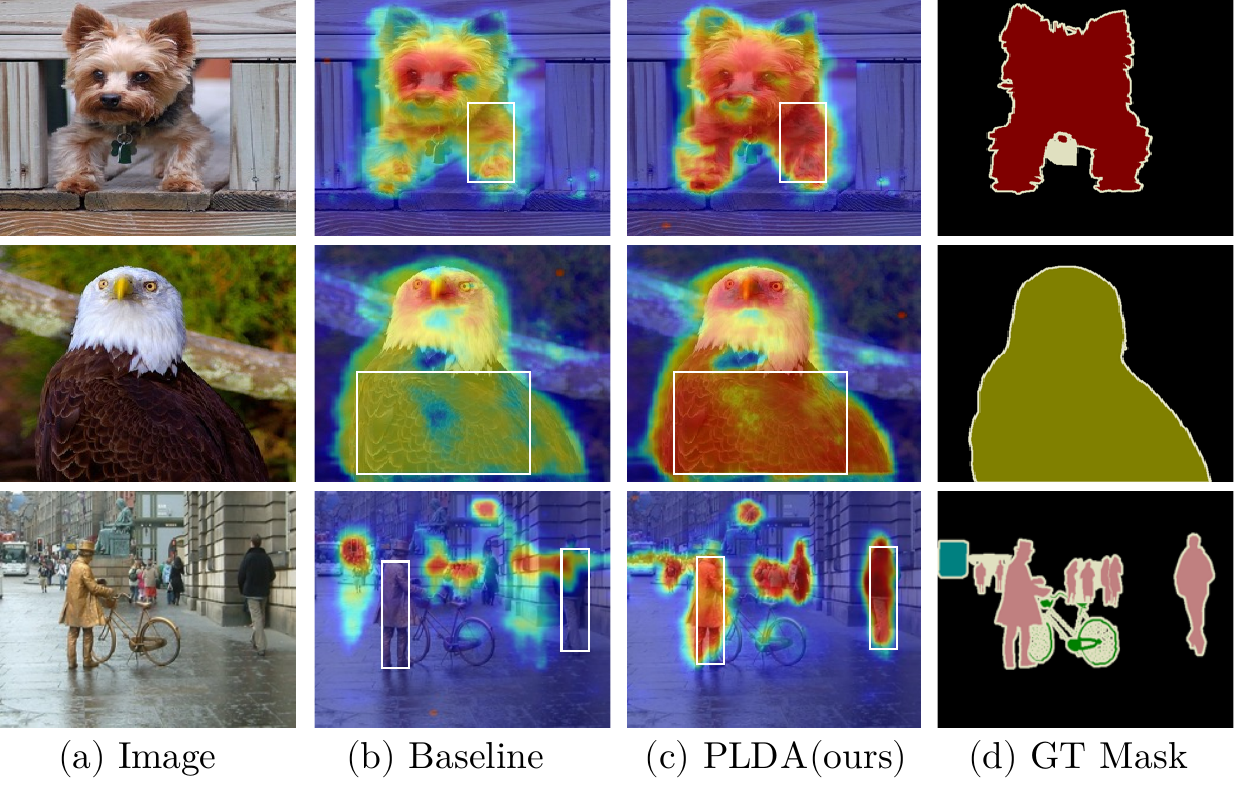}
    \vspace{-10pt}
    \caption{
        \textbf{Comparison} of the CAM between the baseline and our method.
        Our PLDA produces more complete CAM to cover the objects, showing the effectiveness.
        }
    \label{fig:CAM}
    %\vspace{-3pt}
\end{figure}

\subsection{Class Activation Map Evaluation}

\subsubsection{Comparison on PASCAL VOC 2012}

The goal of our method is to boost the classification activation map, \textit{i.e.}, the initial seeds mIoU.
To this end, we conduct experiments to integrate our PLDA into several strong baseline models.
The comparison results between our method with baselines and other recent WSSS methods are shown in \cref{tab:cam-voc}.
For the PSA baseline, we build a stronger version (\textit{i.e.}, $\mbox{PSA}^{+}$) by augmenting it with a PCM module~\cite{wang2020self}, and it achieves 50.3\% mIoU, exceeding the original version by 2.9\%.
On this basis, our PLDA further achieves 53.9\% mIoU on the training set, which even surpasses a recent method ESOL~\cite{li2022expansion}.
Furthermore, we also take the popular SEAM model as the baseline, which has been widely used in many works~\cite{zhang2020causal, su2021context, zhang2021complementary, du2022weakly, lee2022weakly}.
Our PLDA increases SEAM by 3.7\% and 3.8\% mIoU on the training and validation sets, respectively, showing substantial performance improvements.
In \cref{fig:CAM}, we present some qualitative comparisons between SEAM and our PLDA.
It can be seen that our PLDA produces higher quality CAMs than SEAM, especially in non-discriminative object regions highlighted in white bounding boxes.

In addition, we evaluate PLDA on two recently proposed methods, \textit{i.e.}, SIPE and PPC.
Both methods achieve high performances on the PASCAL VOC 2012 dataset.
Nevertheless, our PLDA surpasses SIPE by 1.4\% and 2.1\% mIoU on the training and validation sets, respectively, and performs better than recent ResNet50~\cite{he2016deep} based methods~\cite{lee2021reducing, xie2022cross}.
Based on PPC, our PLDA further achieves 62.5\% and 60.1\% mIoU, which exceeds PPC by 1.0\% and 1.7\%.
Equipped with a simple ResNet38 backbone, our PLDA even exceeds the recent Transformer-based method MCTformer, which employs a powerful DeiT-S~\cite{touvron2021training} as the backbone and achieves a high performance compared to other CNN-based methods.

\begin{table}[t]
    \centering
     
        \caption{\textbf{Evaluation} on semantic segmentation.
        The mIoU (\%) on validation and test sets of PASCAL VOC 2012 is reported.
        ``I.", ``S." and ``L." mean methods using image-level, saliency and language supervision, respectively.}
        \label{tab:voc_seg}
        \vspace{-4pt}
    \renewcommand{\arraystretch}{1.0}
    \scalebox{0.9}{
    \begin{tabular}{l|l|l|c|c|c} \toprule%[1pt] 
        Method                                     &   Publication     & Backbone & Sup.   & Val.   & Test 
        \\ \midrule%[0.8pt]
        SeeNet \cite{hou2018self}                   & NeurIPS'18 & ResNet101  & I.+S.  & 63.1       & 62.8 \\
        DRS \cite{kim2021discriminative}            & AAAI'21    & ResNet101  & I.+S.  & 71.2       & 71.4 \\
        AuxSegNet \cite{xu2021leveraging}           & ICCV'21    & ResNet38  & I.+S.  & 69.0       & 68.6 \\   
        EDAM \cite{wu2021embedded}                  & CVPR'21    & ResNet101  & I.+S.  & 70.9       & 70.6 \\
        EPS \cite{lee2021railroad}                  & CVPR'21    & ResNet101  & I.+S.  & 71.0       & 71.8 \\
        MCIS \cite{wang2022looking}                   & TPAMI'22    & ResNet101  & I.+S.  & 66.2       & 66.9 \\
        RCA \cite{zhou2022regional}                  & CVPR'22    & ResNet38  & I.+S.  & 71.1       & 71.6 \\
        L2G \cite{jiang2022l2g} 
        & CVPR'22    & ResNet101  & I.+S.  & \textbf{72.0}       & 73.0 \\
        Wu et al.~\cite{wu2022adaptive} & ECCV’22 & ResNet101 & I.+S. &71.8  &  \textbf{73.4} \\
        {EPS++ ~\cite{EPSplus10120949}} & TPAMI’23 & ResNet101 & I.+S. &71.7  &  71.9 \\
        
        AffinityNet \cite{ahn2018learning}          & CVPR'18    & ResNet38  & I.     & 61.7       & 63.7 \\
        IRNet \cite{ahn2019weakly}                  & CVPR'19    & ResNet50  & I.     & 63.5       & 64.8 \\
        CDA \cite{su2021context}                    & ICCV'21    & ResNet38  & I.     & 66.1       & 66.8 \\
        CPN \cite{zhang2021complementary} & ICCV'21    & ResNet38  & I.     & 67.8       & 68.5 \\
        OC-CSE \cite{kweon2021unlocking}             & ICCV'21    & ResNet38  & I.     & 68.4       & 68.2 \\
        RIB \cite{lee2021reducing}             & NeurIPS'21    & ResNet101  & I.     & 68.3       & 68.6 \\
        AdvCAM \cite{lee2022anti}                   & TPAMI'22    & ResNet101  & I.     & 68.1       & 68.0 \\   
        ReCAM \cite{chen2022class} & CVPR'22    & ResNet101  & I.  & 68.5       & 68.4 \\
        CLIMS \cite{xie2022cross} & CVPR'22    & ResNet101  & I.+L.  & 69.3      & 68.7 \\
        ESOL \cite{li2022expansion}                  & NeurIPS'22    & ResNet101  & I.  & 69.9       & 69.3 \\

{OCR ~\cite{cheng2023out}} & ICCV'23    & ResNet38  & I. & 67.8       & 68.4  \\

 {FPR \cite{chen2023fpr}} & ICCV'23    & ResNet38  & I.  & 70.0       & 70.6 \\
    
        {ASDT \cite{zhang2023weakly}} & TIP'23    & ResNet38  & I.  & 69.7       & 70.1 \\

        {BAS~\cite{zhai2024background}}  & IJCV'24    & ResNet101  & I.  & 69.6       &  69.9 \\

        \midrule
        SEAM \cite{wang2020self}                  & CVPR'20    & ResNet38  & I.  & 64.5       &  65.7 \\
        \cellcolor{tabblue!20}\ \ \ \ +\ PLDA                  & \cellcolor{tabblue!20}--    & \cellcolor{tabblue!20}ResNet38  & \cellcolor{tabblue!20}I.  & \cellcolor{tabblue!20}{68.7}  &\cellcolor{tabblue!20}{69.7}  \\
        SIPE \cite{chen2022self}                  & CVPR'22   & ResNet38  & I.  & 68.2       &  69.5 \\
        \cellcolor{tabblue!20}\ \ \ \ +\ PLDA                 & \cellcolor{tabblue!20}--   & \cellcolor{tabblue!20}ResNet38  & \cellcolor{tabblue!20}I.  & \cellcolor{tabblue!20}{68.8}       & \cellcolor{tabblue!20}{69.7} \\
        SIPE \cite{chen2022self}                  & CVPR'22   & ResNet101  & I.  & 68.8       & 69.7 \\
        \cellcolor{tabblue!20}\ \ \ \ +\ PLDA                 & \cellcolor{tabblue!20}--   & \cellcolor{tabblue!20}ResNet101  & \cellcolor{tabblue!20}I.  & \cellcolor{tabblue!20}{69.6}       & \cellcolor{tabblue!20}{71.5} \\
        PPC \cite{du2022weakly}                  & CVPR'22   & ResNet38  & I.  & 67.7       &  67.4 \\
        \cellcolor{tabblue!20}\ \ \ \ +\ PLDA                 & \cellcolor{tabblue!20}--   & \cellcolor{tabblue!20}ResNet38  & \cellcolor{tabblue!20}I.  & \cellcolor{tabblue!20}{69.2}      & \cellcolor{tabblue!20}{69.5}\\
        PPC \cite{du2022weakly}                  & CVPR'22   & ResNet101  & I.  & 67.9       &  69.2  \\
        \cellcolor{tabblue!20}\ \ \ \ +\ PLDA                 & \cellcolor{tabblue!20}--   & \cellcolor{tabblue!20}ResNet101  & \cellcolor{tabblue!20}I.  & \cellcolor{tabblue!20}{69.7}       & \cellcolor{tabblue!20}{71.1} \\
        \bottomrule% [1pt]
        \end{tabular} }%}
        \vspace{-10pt}
\end{table}

\subsubsection{Comparison on MS COCO}

Furthermore, we conduct experiments of our PLDA method on the challenging MS COCO 2014 dataset.
We select SIPE model as the baseline due to its notable performance and efficiency. The comparison between PLDA and SIPE, along with other recent approaches, is detailed in \cref{tab:cam-coco}. The results depicted in the table highlight our PLDA's achievement of a substantial 37.3\% mIoU on the training set, significantly exceeding the baseline by 1.7\%.
In contrast to EPS, which incorporates saliency maps for supplementary guidance, our PLDA demonstrates competitive performance. This result shows the effectiveness of our proposed approach.

\begin{table}[t]
\centering
\small
\caption{
    {
    \textbf{Comparison} with adversarial erasing methods.
    CAM and Seg. indicate the mIoU on PSACAL VOC 2012 for the initial CAM and the final segmentation model.
    ``I." and ``S." mean methods using image-level and saliency supervision, respectively.
    }
}
\label{tab:comparison_AE}
\vspace{3pt}
\scalebox{0.822}{\begin{tabular}{l|l|l|c|c|c}
\toprule[1pt]
Method & Publication & Backbone & Sup. &CAM  & Seg. \\ \midrule[0.8pt]
AE-PSL~\cite{wei2017object}    & CVPR'17     & VGG16 & I. & - & 55.0 \\
SeeNet~\cite{hou2018self}    & NeurIPS'18     & ResNet101 & I.+S. &  - & 63.1 \\
GAIN~\cite{li2018tell} & CVPR'18   & VGG16 & I. & - & 55.3  \\
OC-CSE~\cite{kweon2021unlocking}    & ICCV'21     & ResNet38  & I. & 56.0 & 68.4  \\
AEFT~\cite{yoon2022adversarial}    & ECCV'22     & ResNet38   & I. & 56.0 & - \\

\cellcolor{tabblue!20}PLDA (\textit{w/} SEAM)   & \cellcolor{tabblue!20}--          & \cellcolor{tabblue!20}ResNet38  & \cellcolor{tabblue!20}I. & \cellcolor{tabblue!20}{59.1} & \cellcolor{tabblue!20}{68.7} \\

\cellcolor{tabblue!20}PLDA (\textit{w/} PPC)   & \cellcolor{tabblue!20}--          & \cellcolor{tabblue!20}ResNet38 & \cellcolor{tabblue!20}I. & \cellcolor{tabblue!20}\textbf{62.5} & \cellcolor{tabblue!20}\textbf{69.2} \\
\bottomrule[1pt]
\end{tabular}}
\end{table}

{
\subsubsection{Comparison with AE Methods.}
Adversarial erasing methods represent a class of techniques to improve the completeness of CAM.
Considering that our PLDA method uses AE to mine the target domain, we present additional comprehensive comparisons between PLDA and state-of-the-art AE-based methods, as shown in \cref{tab:comparison_AE}.
Specifically, AE-PSL~\cite{wei2017object}, SeeNet~\cite{hou2018self}, and GAIN~\cite{li2018tell} are among the early AE methods, which lay behind our proposed PLDA method by large margins.
A recent method OC-CSE \cite{kweon2021unlocking} proposes a class-specific erasing strategy, which stands for one of the most advanced AE methods.
Note that the most fair comparison should focus on the performance of CAM, as the core of these methods is to generate more complete CAM, rather than downstream CAM refinement and segmentation training.
Compared to OC-CSE, our PLDA on top of SEAM surpasses it by 3.1\% in CAM performance.
Moreover, our PLDA exceeds AEFT \cite{yoon2022adversarial} by a large margin of 6.5\% in CAM performance, demonstrating a clear superiority.
It should also be emphasized that our work only considers AE as a component for domain assignment, rather than proposing a new AE method. 
On the contrary, exploring more advanced domain assignment methods in PLDA is a possible research direction in the future.
}
\vspace{-2em}

\subsection{Semantic Segmentation Evaluation}

\subsubsection{Comparison on PASCAL VOC 2012}
The common practice in WSSS is first generating the pseudo masks from CAM and then train a semantic segmentation model.
In this subsection, we utilize the pseudo masks generated by PLDA to train a DeepLab-LargeFoV (also known as DeepLabV1)~\cite{liang2015semantic} and compare the performance with several baseline methods.
In accordance with common practices, we adopt the widely used IRN \cite{ahn2019weakly} to refine CAMs, which then serve as pseudo ground truths for the segmentation training process.

\cref{tab:voc_seg} presents a comprehensive comparison of results on both the PASCAL VOC 2012 validation and test sets, highlighting the performance of our proposed PLDA method.

It is important to note that the evaluation on the test set is conducted through the official evaluation server, ensuring the credibility of the results.
It can be seen that our PLDA demonstrates significant improvements over the SEAM baseline, with segmentation mIoU values increasing from 64.5\% to 68.7\% on the validation set and from 65.7\% to 69.7\% on the test set. 
Impressively, even compared to recent SEAM-based techniques like CDA and CPN, our PLDA exhibits remarkable advancements.
Moreover, when integrated with the SIPE framework, our PLDA consistently enhances performance using ResNet38 and ResNet101 backbones. 
Notably, with ResNet101, our PLDA achieves a remarkable 71.1\% mIoU on the test set, surpassing SIPE by 1.8\%. 
It is worth noting that our PLDA's performance remains competitive compared to other saliency-based models such as EPS.
Furthermore, our PLDA method significantly improves the performance of the ResNet38-based PPC model by 1.5\% mIoU and 2.1\% mIoU on the validation and test sets, respectively. 
The advantage of PLDA becomes even more pronounced when employed with the ResNet101 backbone, reaching mIoU values of 69.8\% and 71.1\% on the validation and test sets, respectively.
These compelling experimental results underscore the effectiveness of our proposed PLDA method on the PASCAL VOC 2012 dataset, promising enhanced performance and competitiveness in the WSSS task.

\begin{table}[t]
    \centering
    %\scriptsize
     
    \renewcommand{\arraystretch}{1.0}
    \caption{\textbf{Evaluation} on semantic segmentation.
    The mIoU (\%) on the validation set of MS COCO 2014 is reported.
    ``I." and ``S." mean methods using image-level and saliency supervision, respectively.
    {* means the single-stage method.}
    }\label{tab:coco_seg}
    \vspace{-4pt}
    \scalebox{0.9}{
    \begin{tabular}{l|l|l|c|c} \toprule%[1pt]
        Method                                      & Publication             & Backbone & Sup.   & Validation    \\ \midrule%[0.8pt]       
        ADL \cite{choe2020attention}                  & TPAMI'20  &   VGG16   & I.+S.  & 30.8       \\
        EPS \cite{lee2021railroad}                  & CVPR'21  &   VGG16   & I.+S.  & 35.7        \\
        AuxSegNet \cite{xu2021leveraging}  & ICCV’21  & ResNet38 & I.+S. & 33.9  \\
        RCA \cite{zhou2022regional}              & CVPR'22       &   VGG16   & I.+S.  &  36.8           \\
        L2G \cite{jiang2022l2g}              & CVPR'22       &   ResNet101   & I.+S.  &  44.2           \\
        Wu et al.~\cite{wu2022adaptive} & ECCV'22 &VGG16 & I.+S.  & 38.6 \\
          {EPS++~\cite{EPSplus10120949} }& TPAMI'23 &ResNet101 & I.+S.  & 42.4 \\ 
        
        IRN \cite{ahn2019weakly}                 & CVPR'19  &   ResNet50   & I.  & 32.6        \\ 
        CONTA \cite{zhang2020causal}                & NeurIPS'20  &   ResNet101   & I.  & 33.4        \\ 
       OC-CSE \cite{kweon2021unlocking}                  & ICCV'21  &   ResNet38   & I.  & 36.4        \\
        PMM \cite{li2021pseudo}                  & ICCV'21  &   ScaleNet101   & I.  & 40.2        \\
        RIB \cite{lee2021reducing}                  & NeurIPS'21  &   ResNet101   & I.  & 43.8        \\
        URN \cite{li2022uncertainty}                  & AAAI'22  &   Res2Net-101   & I.  & 41.5        \\
        MCTformer \cite{xie2022cross} & CVPR'22 &   ResNet38   & I.  & 42.0        \\ 
        ESOL \cite{li2022expansion} & NeurIPS'22 &   ResNet101   & I.  & 42.6        \\

    {MCIC ~\cite{MCIC9873854}} & TPAMI'22    & VGG16  & I. & 30.2       \\
        
        {USAGE \cite{peng2023usage}} & ICCV'23 &   ResNet38   & I.  & 42.7

        \\
         {OCR \cite{cheng2023out}} & CVPR'23 &   ResNet38   & I.  & 42.5 \\   
        
        {MDBA* \cite{chen2023multi}} & TIP'23 &   ResNet101   & I.  & 37.8  \\        
        \midrule
        SIPE \cite{chen2022self}                  & CVPR'22  &   ResNet38   & I.  & 43.6        \\ 
        \cellcolor{tabblue!20}\ \ \ \ +\ PLDA                   & \cellcolor{tabblue!20}--  &   \cellcolor{tabblue!20}ResNet38   & \cellcolor{tabblue!20}I.  & \cellcolor{tabblue!20}{43.9}
        \\ 
        SIPE \cite{chen2022self}                  & CVPR'22  &   ResNet101   & I.  & 40.6
        \\ 
        \cellcolor{tabblue!20}\ \ \ \ +\ PLDA                   & \cellcolor{tabblue!20}--  &   \cellcolor{tabblue!20}ResNet101   & \cellcolor{tabblue!20}I.  & \textbf{\cellcolor{tabblue!20}}\textbf{44.7}
        \\ 
        \bottomrule%[1pt]
        \end{tabular} }%}
        \vspace{-5pt}
\end{table}

\subsubsection{Comparison on MS COCO}

Furthermore, we extend our evaluation to the challenging MS COCO dataset, with the obtained results being showed in \cref{tab:coco_seg}. 
To ensure a fair evaluation, we seamlessly integrate our PLDA approach into the SIPE framework, renowned for its simplicity and efficacy on MS COCO.
Consistent with SIPE's setup, we opt for the use of dense CRF for CAM refinement, replacing the previously employed IRN.
Remarkably, our PLDA achieves a commendable 43.9\% mIoU with the ResNet38 backbone. Additionally, when equipped with the ResNet101 architecture, our PLDA attains the pinnacle of performance at 44.7\% mIoU, exceeding SIPE by 4.1\% mIoU. This result not only signifies a substantial improvement over the baseline, but it also eclipses the achievements of numerous contemporary saliency-based models, exemplified by L2G.
These results eloquently demonstrate the effectiveness of our PLDA method even in the face of challenging scenarios posed by MS COCO.

\begin{table}[t]
\centering
\caption{\textbf{Ablation} of each component in our PLDA method.}
\label{tab:ablation_component}
% \vspace{-pt}
\setlength{\tabcolsep}{7pt}
\scalebox{0.95}{
 {
\begin{tabular}{ccccc}
\toprule
Baseline & $\mathcal{L}_{uda}$ & $\mathcal{L}_{cps_s}$ & $\mathcal{L}_{cps_t}$  & mIoU (\%) \\ \midrule
 $\checkmark$       &         &         &         & 55.4 \\
 $\checkmark$       & $\checkmark$       &         &         & 57.1 \\
 $\checkmark$       &         & $\checkmark$       &         & 56.1 \\
 $\checkmark$       &         &       & $\checkmark$        & 56.4 \\
 $\checkmark$      & $\checkmark$       & $\checkmark$       &         & 58.5 \\
\cellcolor{tabblue!20}$\checkmark$       & \cellcolor{tabblue!20}$\checkmark$       & \cellcolor{tabblue!20}$\checkmark$       & \cellcolor{tabblue!20}$\checkmark$       & \cellcolor{tabblue!20}\textbf{59.1} \\
\bottomrule
\end{tabular}}}
\vspace{-8pt}
\end{table}

\begin{figure}[t]
    \centering
    \includegraphics[width=0.48\textwidth]{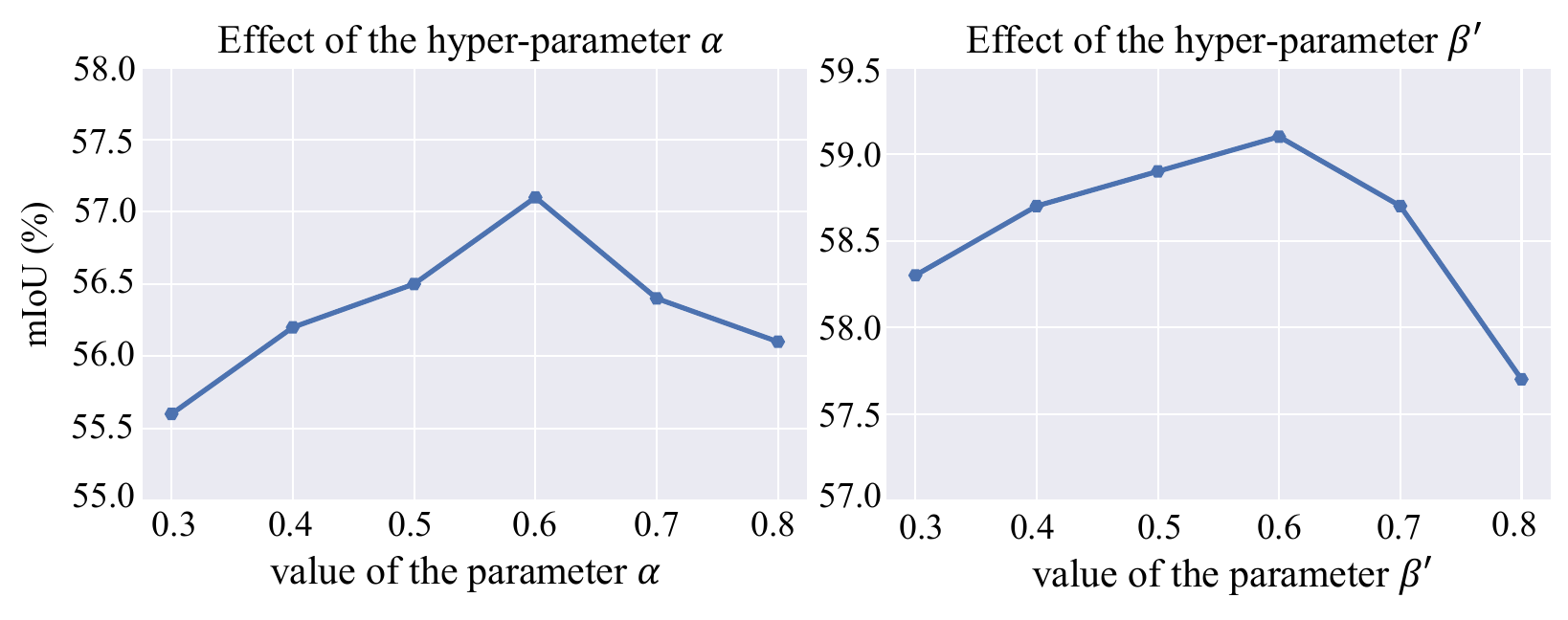}
    \caption{{\textbf{Ablation} of hyper-parameters $\alpha$ and $\beta'$ in our PLDA method.}}
    \label{alpha_beta}
\end{figure}

\begin{table}[t]
\centering
\caption{\textbf{Comparison} of domain adaptation methods.}
\label{tab:comp_uda_method}
\vspace{-3pt}
\setlength{\tabcolsep}{7pt}
\scalebox{1.0}{ {\begin{tabular}{lcc}
\toprule
Method  & Disentanglement & mIoU (\%)     \\ \midrule
Baseline    & --              & 55.4     \\
MMD \cite{dziugaite2015training}    & \ding{55}                  & 56.1 \\
DANN \cite{ganin2015unsupervised}  & \ding{55}                     & 56.2     \\
MADA \cite{pei2018multi}     & $\checkmark$          & 56.6 \\
\cellcolor{tabblue!20}\cref{eq:multiuda} (Ours)     & \cellcolor{tabblue!20}$\checkmark$         & \cellcolor{tabblue!20}\textbf{57.1}    \\\bottomrule
\end{tabular}}}
\end{table}

\subsection{Ablation Study} \label{sec:exp_ablation}

\subsubsection{Component Analysis of PLDA}

As delineated earlier, our PLDA methodology encompasses a trio of integral components, namely the unsupervised domain adaptation loss, and the two confident pseudo-supervision losses applied to the source and target domains correspondingly.
We initiate our investigation with a thorough series of ablation experiments, meticulously assessing the efficacy of each individual component within our PLDA framework. The outcomes of these comprehensive analyses are succinctly depicted in \cref{tab:ablation_component}.

Obviously, the $\mathcal{L}_{uda}$ term, as expressed in \cref{eq:multiuda}, elicits a discernible improvement in the mIoU metric, elevating the baseline SEAM model from 55.4\% to a commendable 57.1\%. This enhancement cogently shows both the imperative \textit{necessity} and profound \textit{effectiveness} inherent within our devised alignment strategy.
Concomitantly, the $\mathcal{L}{cps_s}$ and $\mathcal{L}{cps_t}$ components respectively contribute gains of 0.7\% and 1.0\% mIoU. Notably, the fusion of $\mathcal{L}_{uda}$ and $\mathcal{L}_{cps_s}$ culminates in a cumulative mIoU of 58.5\%, thus validating the \textit{complementary} effect between adversarial domain alignment and the confident pseudo-supervision.
Of paramount significance is the pinnacle achievement of 59.1\% mIoU, seamlessly realized through the collective synergy of all three loss terms. This remarkable accomplishment undeniably outpaces the baseline model's performance by a substantial margin of 3.7\%, thereby proving the effectiveness of our PLDA method.

\subsubsection{Comparison of Domain Adaptation Methods.}

Subsequently, we undertake a comprehensive comparison between our devised domain alignment strategy and several established domain adaptation methodologies, which encompass MMD, DANN, and MADA.
It is noteworthy that these techniques conventionally operate at the image level. To harmonize the evaluation with our pixel-level focus in weakly supervised semantic segmentation, we adeptly extend these approaches to function at the pixel granularity.
Both MMD and DANN (\textit{i.e.}, \cref{eq:uda_loss}) are representatives of the global alignment paradigm, and MADA is a classical category-wise alignment method.

As presented in \cref{tab:comp_uda_method}, our novel approach (\textit{i.e.}, \cref{eq:multiuda}) emerges as the definitive leader amongst this array of domain adaptation methodologies. Specifically, our approach distinctly surpasses DANN by 0.9\% and MADA by 0.5\% in terms of mIoU. These results not only confirm the necessity of reducing misalignment between classes, but also firmly establish the ascendancy of our proposed method.

\begin{table}[t]
\centering
\caption{\textbf{Comparison} of domain label assignment methods. The term ``SimpleAssign" refers to the assignment approach previously mentioned, and ``OC" means original classifier proposed in \cite{kweon2021unlocking}.}
\label{tab:comp_assign_methods}
\vspace{-4pt}
\setlength{\tabcolsep}{10pt}
\scalebox{1.05}{ {\begin{tabular}{lc}
\toprule
Method          & mIoU (\%) \\
\midrule
Baseline          & 55.4      \\
SimpleAssign      & 56.5      \\
MaskAssign-OC    & 56.6      \\
\cellcolor{tabblue!20}MaskAssign (Ours) & \cellcolor{tabblue!20}\textbf{57.1}      \\
\bottomrule
\end{tabular}}}
\end{table}

\subsubsection{Comparison of Domain Assignment Strategies}
We proceed to evaluate our proposed masked domain label assignment strategy, subjecting it to comparison against alternative methods.

As delineated in \cref{tab:comp_assign_methods}, the direct implementation of the SimpleAssign strategy employing dual thresholds on the class activation maps yields a modest 56.5\% mIoU value. In stark contrast, our MaskAssign strategy surges ahead, attaining a superior 57.1\% mIoU.
Moreover, we extend our evaluation to embrace the ordinary classifier (OC) approach, as espoused in \cite{kweon2021unlocking}, for conducting masked assignments of target domain pixels. While achieving a commendable 56.6\% mIoU, this approach still falls short of our proposed strategy.
These comprehensive experimental results show the superiority of our MaskAssign strategy, affirming its efficacy in comparison to other alternatives.

\subsubsection{Sensitivity to Domain Assignment}

In a pursuit to validate the robustness of our proposed PLDA method, we embark on an exploration of its sensitivity to domain assignments.
Specifically, we perform PLDA alignment on different source and target domain data selected under various thresholds (\textit{i.e.}, $\alpha$), and the results are shown in \cref{alpha_beta}(a).
Compared with the baseline model, PLDA achieves significant improvements even at extreme settings (\textit{e.g.} $\alpha$=0.8, where only a small fraction of pixels are assigned as the target domain), thus demonstrating the robustness of PLDA. This remarkable resilience across the spectrum of threshold values definitively underscores the inherent robustness of our PLDA framework.

\subsubsection{Effect of Hyper-parameters}

In this subsection, we analyze the impact of hyper-parameters on the performance of our PLDA model. 
Specifically, we investigate the effects of two key hyper-parameters: $\alpha$ and $\beta'$. These hyper-parameters play a crucial role in determining the allocation of pixels between the source and target domains, as well as influencing the balance between source and target losses (as illustrated in Eq. 2 and Eq. 3 respectively). Through empirical evaluations, we determine the optimal values for these hyper-parameters.

We begin by evaluating the performance of different $\alpha$ values. From the results presented in \cref{alpha_beta}(a), it is evident that the best performance is achieved when $\alpha$ is set to 0.6. Consequently, we select $\alpha$ = 0.6 as the value for our experiments.
Moving forward, we proceed to assess the impact of $\beta'$. By examining the outcomes presented in \cref{alpha_beta}(b), it is observed that the optimal result, with a mIoU of 59.1\%, is attained when $\beta'$ is set to 0.6. 

\begin{table*}[t]
\centering
 
\caption{\textbf{Comparison} of category performance on PASCAL VOC 2012 validation set.
``R38" and ``R101" mean methods using ResNet38 and ResNet101 as the backbone network, respectively.
$^*$ means our training results using the official pseudo-masks provided by the author.
The best result of each category is marked in bold.
Our PLDA improves the baseline methods by large margins on most categories, which demonstrates the effectiveness.
}
%\vspace{3pt}
\label{tab:category_voc_val}
\renewcommand{\arraystretch}{1.0}%{1.05}
\scalebox{0.9}{\setlength{\tabcolsep}{1.0mm}{
\begin{tabular}{l|ccccccccccccccccccccc|c}
\toprule[1pt]
Method      & bkg           & aero          & bike          & bird          & boat          & bottle        & bus           & car           & cat           & chair         & cow           & table         & dog           & horse         & mbk           & person        & plant         & sheep         & sofa          & train         & tv            & mean          \\ \midrule[0.8pt]
SEC~\cite{kolesnikov2016seed}         & 82.4          & 62.9          & 26.4          & 61.6          & 27.6          & 38.1          & 66.6          & 62.7          & 75.2          & 22.1          & 53.5          & 28.3          & 65.8          & 57.8          & 62.3          & 52.5          & 32.5          & 62.6          & 32.1          & 45.4          & 45.3          & 50.7          \\
AdvErasing~\cite{wei2017object}  & 83.4          & 71.1          & 30.5          & 72.9          & 41.6          & 55.9          & 63.1          & 60.2          & 74.0          & 18.0          & 66.5          & 32.4          & 71.7          & 56.3          & 64.8          & 52.4          & 37.4          & 69.1          & 31.4          & 58.9          & 43.9          & 55.0          \\
CIAN~\cite{fan2020cianCIAN}        & 88.2          & 79.5          & 32.6          & 75.7          & 56.8          & 72.1          & 85.3          & 72.9          & 81.7          & 27.6          & 73.3          & 39.8          & 76.4          & 77.0          & 74.9          & 66.8          & 46.6          & 81.0          & 29.1          & 60.4          & 53.3          & 64.3          \\
BES~\cite{chen2020weakly}         & 88.9          & 74.1          & 29.8          & 81.3          & 53.3          & 69.9          & \textbf{89.4}          & 79.8          & 84.2          & 27.9          & 76.9          & 46.6          & 78.8          & 75.9          & 72.2          & 70.4          & 50.8          & 79.4          & 39.9          & 65.3          & 44.8          & 65.7          \\
ESC-Net~\cite{su2021context}     & 89.8          & 68.4          & 33.4          & 85.6          & 48.6          & \textbf{72.2}          & 87.4          & 78.1          & 86.8          & 33.0          & 77.5          & 41.6          & 81.7          & 76.9          & 75.4          & 75.6          & 46.2          & 80.7          & 43.9          & 59.8          & 56.3 & 66.6          \\ 
CDA~\cite{su2021context}&89.1& 69.7& 34.5& 86.4& 41.3& 69.2& 81.3 &79.5& 82.1 &31.1& 78.3& 50.8 &80.6 &76.1& 72.2& 77.6 &48.8 &81.2& 42.5& 60.6& 54.3& 66.1 \\
OC-CSE~\cite{kweon2021unlocking}&90.2& 82.9 & 35.1& 86.8 &59.4 &70.6& 82.5& 78.1& 87.4 &30.1& 79.4& 45.9& 83.1& \textbf{83.4}& 75.7& 73.4 &48.1& \textbf{89.3} &42.7& 60.4 &52.3 &68.4\\
AdvCAM~\cite{lee2022anti} &90.0& 79.8& 34.1 &82.6 &\textbf{63.3}& 70.5& \textbf{89.4}& 76.0 &87.3& 31.4& 81.3 &33.1 &82.5& 80.8& 74.0& 72.9& 50.3& 82.3 &42.2& \textbf{74.1} &52.9& 68.1 \\
\midrule
SEAM~\cite{wang2020self}$_{\text{R}38}$        & 88.8          & 68.5          & 33.3          & 85.7          & 40.4          & 67.3          & 78.9          & 76.3          & 81.9          & 29.1          & 75.5          & 48.1          & 79.9          & 73.8          & 71.4          & 75.2          & 48.9          & 79.8          & 40.9          & 58.2          & 53.0          & 64.5          \\
\rowcolor{tabblue!20}\ \ +PLDA & \textbf{91.4} & 83.4 & \textbf{38.6} & 86.3 & 52.3 & 65.5 & 85.6 & 82.1 & 87.9 & 31.1 & 68.1 & \textbf{51.5} & 83.3 & 74.4 & 75.5 & 81.5 & 44.6 & 81.9 & 46.4 & 66.7 & \textbf{64.5} & 68.7
         \\
SIPE~\cite{chen2022self}$_{\text{R}38}$ & -- &-- &-- &-- &-- &-- &-- &-- &-- &-- &-- &-- &-- &-- &-- &-- &-- &-- &-- &-- & --  & 68.2          \\
\rowcolor{tabblue!20}\ \ +PLDA &90.3 &82.8 &36.7 &\textbf{91.3} &48.9 &71.1 &87.6 &80.7 &89.1 &25.4 &79.8 &49.1 &86.5 &75.8 &74.3 &73.0 &50.7 &87.0 &\textbf{46.6} &61.8 &55.7 &68.8           \\
SIPE~\cite{chen2022self}$_{\text{R}101}$ & -- &-- &-- &-- &-- &-- &-- &-- &-- &-- &-- &-- &-- &-- &-- &-- &-- &-- &-- &-- &--  & 68.8          \\
\rowcolor{tabblue!20}\ \ +PLDA & 90.3 &83.9 &37.5 &90.3 &49.5 &68.2 &88.3 &80.8 &90.2 &27.9 &\textbf{85.5} &48.9 &\textbf{89.3} &79.4 &76.0 &70.7 &\textbf{56.8} &86.0 &43.1 &63.4 &56.0 &69.6          \\
PPC~\cite{du2022weakly}$_{\text{R}38}$ & 90.0          & 81.9          & 34.2          & 89.2 & 38.6          & \textbf{72.2}          & 88.1          & 81.7          & 89.0          & 30.8          & 70.0          & 50.7 & 83.8          & 75.4          & 76.3          & 75.8          & 49.2          & 84.7          & 43.7 & 58.7          & 53.5          & 67.7          \\
\rowcolor{tabblue!20}\ \ +PLDA & \textbf{91.4} &85.2 &37.8 &89.8 &55.8 &68.8 &86.6 &80.6 &89.8 &31.9 &70.6 &50.2 &84.6 &77.1 &76.4 &82.0 &47.6 &83.3 &\textbf{46.6} &57.1 &59.5 &69.2 \\ 
PPC$^*$~\cite{du2022weakly}$_{\text{R}101}$ & 89.9 &85.3 &34.1 &90.3 &39.6 &69.7 &86.9 &80.6 &89.6 &33.5 &70.2 &49.5 &84.4 &82.1 &73.7 &75.1 &50.7 &85.0 &41.6 &\textbf{59.1} &54.1 &67.9       \\
\rowcolor{tabblue!20}\ \ +PLDA & 91.2 &\textbf{87.8} &37.7 &90.9 &51.9 &69.9 &85.6 &\textbf{82.6} &\textbf{90.7} &\textbf{34.5} &70.0 &48.7 &85.9 &80.6 &\textbf{77.6} &\textbf{82.9} &50.6 &84.7 &43.0 &54.8 &63.2 &\textbf{69.7} \\
\bottomrule[1pt]
\end{tabular}}}
\end{table*}

\begin{table*}[t]
\centering
 
\caption{\textbf{Comparison} of category performance on PASCAL VOC 2012 test set.
``R38" and ``R101" mean methods using ResNet38 and ResNet101 as the backbone network, respectively.
$^*$ means our training results using the official pseudo-masks provided by the author.
The best result of each category is marked in bold.
The table footnotes present the anonymous links on the official evaluation server.
Our PLDA improves the baseline methods by large margins on most categories, which demonstrates the effectiveness.
}
%\vspace{3pt}
\label{tab:category_voc_test}
\begin{threeparttable}
\renewcommand{\arraystretch}{1.0}%{1.05}
\scalebox{0.9}{\setlength{\tabcolsep}{1.0mm}{
\begin{tabular}{l|ccccccccccccccccccccc|c}
\toprule[1pt]
Method      & bkg           & aero          & bike          & bird          & boat          & bottle        & bus           & car           & cat           & chair         & cow           & table         & dog           & horse         & mbk           & person        & plant         & sheep         & sofa          & train         & tv            & mean          \\ \midrule[0.8pt]
CPN~\cite{zhang2021complementary}&90.4&79.8&32.9& {85.8}&52.9&66.4&87.2&81.4&87.6&28.2&79.7&50.2&82.9&80.4&78.9&70.6&51.2&83.4&55.4&\textbf{68.5}&44.6&68.5 \\
AdvCAM~\cite{lee2022anti}& 90.1 &81.2 &33.6& 80.4 &52.4& 66.6& 87.1& 80.5& 87.2& 28.9& 80.1& 38.5& 84.0 &83.0& 79.5 & 71.9& 47.5&80.8&\textbf{59.1}& 65.4 &49.7 &68.0 \\
\midrule
SEAM~\cite{wang2020self}$_{\text{R}38}$        & --          & --         & --          & --      
& --          & --          & --          & 
--          & --          & --          & 
--          & --          & --          & 
--          & --          & --          & 
--          & --          & --          & 
--          & --          & 65.7          \\
\rowcolor{tabblue!20}\ \ +PLDA & \textbf{91.7} &81.3 &38.2 &89.0 &47.5 &63.5 &83.8 &82.4 &86.2 &27.1 &76.5 &60.4 &79.3 &79.4 &83.3 &79.7 &41.2 &85.5 &56.8 &\textbf{68.5} &\textbf{63.1} &69.7         \\
SIPE~\cite{chen2022self}$_{\text{R}38}$  &      90.5 &84.0 &36.0 &89.5 &\textbf{60.8} &62.1 &87.2 &78.7 &80.3 &\textbf{32.9} &76.5 &57.9 &77.5 &79.1 &80.4 &73.4 &\textbf{59.4} &83.1 &57.0 &62.0 &50.4 &69.5       \\
\rowcolor{tabblue!20}\ \ +PLDA & 90.3 &84.9 &37.3 &89.0 &44.4 &67.4 &86.0 &82.3 &88.8 &24.3 &82.2 &56.7 &85.2 &82.4 &81.2 &72.2 &58.2 &86.8 &57.3 &52.3 &54.1 &69.7 \\
SIPE~\cite{chen2022self}$_{\text{R}101}$         & 90.3 &84.3 &35.2 &87.3 &59.0 &57.4 &\textbf{88.7} &79.9 &83.3 &31.5 &83.3 &56.7 &82.8 &85.4 &79.1 &70.3 &51.7 &85.7 &57.8 &61.9 &52.4 &69.7          \\
\rowcolor{tabblue!20}\ \ +PLDA &  90.7 &83.8 &38.2 &\textbf{92.6} &47.9 &\textbf{71.1} &88.6 &82.4 &89.2 &25.0 &\textbf{87.9} &58.3 &\textbf{86.6} &\textbf{87.4} &\textbf{85.1} &71.0 &59.2 &86.4 &57.7 &56.6 &56.7 &\textbf{71.5} \\
PPC~\cite{du2022weakly}$_{\text{R}38}$         & 90.1 &83.8 &34.2 &87.6 &32.7 &66.2 &88.0 &83.8 &88.0 &28.6 &73.1 &60.0 &80.9 &79.2 &80.6 &74.1 &45.7 &81.0 &52.9 &52.9 &51.1 &67.4         \\
\rowcolor{tabblue!20}\ \ +PLDA &  91.2 &85.8 &37.8 &89.8 &48.5 &66.9 &85.3 &81.8 &88.6 &28.7 &74.7 &61.2 &82.6 &81.2 &81.6 &79.1 &43.9 &85.6 &57.1 &50.2 &58.3 &69.5  \\ 
PPC$^*$~\cite{du2022weakly}$_{\text{R}101}$         & 90.3 &85.0 &34.9 &91.2 &36.0 &69.2 &86.4 &\textbf{86.7} &89.9 &25.6 &76.7 &\textbf{61.5} &81.5 &86.5 &81.2 &74.7 &50.1 &86.4 &53.2 &54.1 &52.3 &69.2         \\
\rowcolor{tabblue!20}\ \ +PLDA &  91.5 &\textbf{86.6} &\textbf{38.7} &92.5 &50.8 &70.1 &87.9 &85.3 &\textbf{91.5} &25.7 &78.5 &61.0 &85.4 &87.3 &83.1 &\textbf{80.7} &46.5 &\textbf{86.9} &54.7 &47.9 &61.3 &71.1 \\ 
\bottomrule[1pt]
\end{tabular}}}
% \begin{tablenotes}
% \item$^1$\href{http://host.robots.ox.ac.uk:8080/anonymous/NZW0KI.html}{http://host.robots.ox.ac.uk:8080/anonymous/NZW0KI.html}
% \item$^2$\href{http://host.robots.ox.ac.uk:8080/anonymous/PNOZY1.html}{http://host.robots.ox.ac.uk:8080/anonymous/PNOZY1.html} 
% \item$^3$\href{http://host.robots.ox.ac.uk:8080/anonymous/KVSK2A.html}{http://host.robots.ox.ac.uk:8080/anonymous/KVSK2A.html}
% \item$^4$\href{http://host.robots.ox.ac.uk:8080/anonymous/PZANKB.html}{http://host.robots.ox.ac.uk:8080/anonymous/PZANKB.html}
% \item$^5$\href{http://host.robots.ox.ac.uk:8080/anonymous/AWRJ05.html}{http://host.robots.ox.ac.uk:8080/anonymous/AWRJ05.html}
% \item$^6$\href{http://host.robots.ox.ac.uk:8080/anonymous/GXY7VD.html}{http://host.robots.ox.ac.uk:8080/anonymous/GXY7VD.html}
% \end{tablenotes}
\end{threeparttable}
\end{table*}
\vspace{-1em}

\begin{table}[t]
\centering
 
\caption{\textbf{Comparison} of category performance on MS COCO 2014 validation set. “R38” and “R101” mean methods using ResNet38 and ResNet101 as the backbone network, respectively. The best result of each category is marked in bold.
Our PLDA method achieves the best performance on most categories.
}
\label{tab:category_coco}
\centering
\renewcommand{\arraystretch}{1.0}
\scalebox{0.72}{\setlength{\tabcolsep}{1.3mm}{
\begin{tabular}{l|ccc>{\columncolor{tabblue!20}}c>{\columncolor{tabblue!20}}c|l|ccc>{\columncolor{tabblue!20}}c>{\columncolor{tabblue!20}}c}
\toprule[1pt]
\rotatebox{90}{Category} & \rotatebox{90}{Luo et al.~\cite{luo2020learning}} & \rotatebox{90}{AuxSegNet~\cite{xu2021leveraging}} & \rotatebox{90}{MCTformer~\cite{xu2022multi}} &
\rotatebox{90}{PLDA (Ours)$_{\text{R}38}$}  &
\rotatebox{90}{PLDA (Ours)$_{\text{R}101}$}
&
\rotatebox{90}{Category} & \rotatebox{90}{Luo et al.~\cite{luo2020learning}} & \rotatebox{90}{AuxSegNet~\cite{xu2021leveraging}} & \rotatebox{90}{MCTformer~\cite{xu2022multi}}
&
\rotatebox{90}{PLDA (Ours)$_{\text{R}38}$}  &
\rotatebox{90}{PLDA (Ours)$_{\text{R}101}$}
         \\ 
\midrule[0.8pt]
background &73.9&82.0&{82.4} & \textbf{83.0}   & 82.7
                           
&wine class&27.2&{32.1}&27.0 & \textbf{43.4}        & 40.7
                                  \\
person &48.7&{65.4}&62.6 & \textbf{56.4}     & 55.9
                                    
&cup &21.7&{29.3}&29.0  & \textbf{37.9}    & 35.5
                                           \\
bicycle &45.0&43.0&{47.4} & \textbf{52.1}      & 51.3
                                       
&fork &0.0&5.4&{13.9}  & 17.8           & \textbf{21.9}
                                     \\
car &31.5&34.5& \textbf{47.2} & 37.8             & 40.1
                                    
& knife &0.9&1.4&{12.0} & \textbf{17.9}          & 17.0
                                     \\
motorcycle &59.1&{66.2}&63.7 & \textbf{72.6}    & 71.6
                                      
& spoon &0.0&1.4&{6.6}&  9.1            & \textbf{13.7}
                                    \\ 
airplane &26.9&60.3&{64.7} & 69.0          & \textbf{74.8}
                                
& bowl &7.6&19.5&{22.4} & 25.8       & \textbf{26.5}
                                        \\ 
bus &52.4&63.1&{64.5}& \textbf{65.7}  & 58.6 &
                                    
banana&52.0&46.9&{63.2} & 55.9            & \textbf{65.9}
                                   \\
train&42.4&57.3&\textbf{64.5} & 48.7           & 50.7
                           
& apple &28.8&40.4&{44.4} & 42.1      & \textbf{51.7}
                                       \\
truck &36.9&38.9&{44.8} & 42.8          & \textbf{45.6}
                           
& sandwich&37.4&39.4& \textbf{39.7} & 38.4    & 37.6
                                       \\
boat &23.5&30.1&\textbf{42.3} & 41.0               & 35.2
         
& orange &52.0&52.9&{63.0} & 55.5       & \textbf{68.8}
                                      \\
traffic light &13.3&40.4&\textbf{49.9} & 40.9     & 39.9
                          
& broccoli &33.7&36.0&{\textbf{51.2}} & 49.1        & 45.9
                                  \\
fire hydrant &45.1&72.7&{73.2} & \textbf{77.8}     & 67.3
                          
& carrot &29.0&13.9&{40.0} & \textbf{40.8}       & 29.3
                                    \\
stop sign &43.4&40.3&{76.6} & \textbf{76.9}     & 68.8
                             
& hot dog &38.8&46.1&{\textbf{53.0}} & 44.2      & 44.3
                                       \\
parking meter &33.5&59.8&{64.4} & 67.1      & \textbf{76.2}
                         
& pizza &{69.8}&62.0&62.2 & \textbf{65.3}      & 48.4
                                     \\
bench &26.3&16.0&{32.8} & 31.9                & \textbf{36.7}
                                 
& donut &50.8& 43.9&{\textbf{55.7}}  & 53.6          & 53.6
                                      \\
bird &29.9&61.0&{62.6}  & 65.3            & \textbf{72.4}
                    
& cake &37.3&30.6&{47.9}  & \textbf{54.4}        & 48.7
                                     \\
cat &62.1&68.6&{\textbf{78.2}}  & 72.9          & 72.7
                       
& chair &10.7&11.4&{22.8}  & 22.2          & \textbf{24.8}
                                 \\
dog &57.5&66.9&{68.2} & 63.0              & \textbf{69.1}
                    
& couch &9.4& 14.5&{35.0}  & \textbf{36.2}       & 36.1
                                   \\
horse &40.7&55.6&{65.8}  & 64.6            & \textbf{68.3}
                   
& potted plant &{\textbf{21.8}}&2.1&13.5  & 17.4       & 20.2
                                 \\
sheep &54.0&61.4&{70.1}  & 71.2                   & \textbf{72.6}
            
& bed &34.6&20.5&{48.6} &  \textbf{51.9}            & 50.2
                              \\
cow &47.2&60.7&{68.3}  & \textbf{70.5}                   & 68.0
              
& dining table &1.1&9.5&{12.9}  & \textbf{15.7}             & 13.3
                            \\
elephant &64.3&76.1&{\textbf{81.6}} & 80.6                & 81.5
                                
& toilet &43.8&57.8&{63.1} & 66.1               & \textbf{64.7}
                           \\
bear &58.9&73.0&{\textbf{80.1}} & 72.6                & 77.0
                 
& tv &11.5&36.0&{\textbf{47.9}} &  46.0                     & 46.4
                     \\
zebra&60.7&80.8&{83.0}  & 83.0                     & \textbf{84.3}
           
& laptop &37.0&35.2&{49.5} & 51.9              & \textbf{53.0}
                          \\
giraffe &45.1&71.6&{76.9}  & \textbf{79.1}                  & 73.2
            
& mouse &0.0&{13.4}&{13.4} & 7.3              & \textbf{15.0}
                           \\
backpack &0.0&11.3&{14.6}  & 19.9                 & \textbf{30.5}
             
& remote&37.2&23.6&{\textbf{41.9}}  & 29.9               & 38.7
                                           \\
umbrella &46.1&35.0&{61.7} & 64.6                 & \textbf{68.7}
                                   
& keyboard &19.0&17.9&{49.8}  &  \textbf{57.1}                 & 53.2
                                      \\
handbag &0.0&2.2&{4.5}  & \textbf{9.2}                      & 9.0
                              
& cellphone &38.1&49.9&{54.1}  & \textbf{56.9}               & 51.8
                                             \\
tie &15.5&14.7&{\textbf{25.2}}  & 4.8               & 3.4
                                     
& microwave &{\textbf{43.4}}&28.7&38.0  & 35.2                 & 42.3
                                          \\
suitcase &43.6&31.7&{46.8}  & 46.4           & \textbf{47.6}
                                        
& oven &29.2&13.3&{29.9}  & 30.6                    & \textbf{32.6}
                                      \\
frisbee &23.2&1.0&{43.8}  & \textbf{63.2}               & 48.3
                  
& toaster &0.0&0.0 &0.0  & 0.0                       & 0.0
                                                \\
skis &6.5&8.1&{12.8} & \textbf{17.4}                          & 17.2
                             
& sink &{28.5}&21.0&28.0 & 32.0                 & \textbf{34.7}
                                            \\
snowboard &10.9&7.6&{31.4} & \textbf{38.3}                 & 34.6
                                   
& refrigerator &23.8&16.6&{40.1} & \textbf{47.6}           & 36.5
                                                    \\
sports ball &0.6&{28.8}&9.2 & 36.2                   & \textbf{39.8}
                                
& book &26.3&8.7&\textbf{32.2}  & 24.0                  & 27.0
                                              \\
kite &14.0&{27.3}&26.3 & \textbf{41.2}                     & 35.9
                                  
& clock &13.4&34.4&{43.2}  & \textbf{46.6}                 & 43.0
                                                      \\
baseball bat &0.0&{2.2}&0.9 & 21.7                 & \textbf{24.7}
                                                
& vase &{27.1}&25.9&22.6 & 26.4                       & \textbf{31.2}
                                                  \\
baseball globe &0.0&{1.3}&0.7 & \textbf{8.1}                & 5.6
                                                 
& scissors &{37.0}&16.6&32.9  & 38.9                & \textbf{48.2}
                                                       \\
skateboard &7.6&{15.2}&7.8  & 19.7                      & \textbf{30.4}
                                            
& teddy bear &58.9&47.3&{61.9} & \textbf{63.7}                  & 62.3
                                                        \\
surfboard &17.6&17.8&{\textbf{46.5}}  & 45.1                   & 45.3
                                              
& hair drier &0.0&0.0&0.0 & 0.0                 & 0.0
                                                               \\
tennis racket &38.1&{\textbf{47.1}} &1.4 & 22.8                & 26.3
                            
& toothbrush &11.1&1.4&{12.2} & 25.7           & \textbf{43.2}
                                                               \\
bottle &28.4&{33.2}&31.1  & 34.1                                     & \textbf{35.3}
      
& \textbf{mean} &29.9&33.9&{42.0} & {43.9}     & \textbf{44.7}
                                                 \\
\bottomrule[1pt]
\end{tabular}
}}
\end{table}

\subsection{Comparison of Category Performance}\label{sec:exp_categoty}

\subsubsection{Comparison on PASVAL VOC 2012}
\cref{tab:category_voc_val} presents a comprehensive comparison of category performance on the PASCAL VOC 2012 validation dataset. 
The results demonstrate notable advancements achieved by our proposed PLDA approach over baseline methods across various categories. 
In particular, when integrated with a ResNet38 backbone, our PLDA method outperforms the SEAM model, achieving significant improvements of 14.9\%, 11.9\% and 6.7\% in mIoU values for the categories of airplane, boat and bus, respectively. Furthermore, in comparison to the PPC approach, our PLDA method consistently achieves superior performance in most categories, such as person, sofa, and television, when utilizing the ResNet-38 backbone. In addition, an improvement of 7.8\% in mIoU is observed for the category of person of the PPC model in the validation set, demonstrating the efficacy of our PLDA method when coupled with the ResNet-101 backbone.

Similar trends are observed when evaluating performance on the test set. Our PLDA approach consistently surpasses baseline models in a majority of categories. The results displayed in \cref{tab:category_voc_test} highlight the superiority of our PLDA method, particularly with a ResNet-101 backbone, outperforming the SIPE baseline in 15 out of the 21 categories tested. Furthermore, substantial performance gains are evident when comparing our PLDA method with the PPC baseline, particularly in categories such as bird, dog, and horse. These compelling comparisons demonstrate the effectiveness of our proposed PLDA approach in achieving higher category-level results.

\subsubsection{Comparison on MS COCO 2014}
\cref{tab:category_coco} shows the category performance of our proposed PLDA method on MS COCO 2014.
In contrast to AuxSegNet, our PLDA method showcases superior performance across numerous categories.
Particularly noteworthy is its achievement of more than a 10\% improvement in mIoU over AuxSegNet in several categories, including boat, bicycle, umbrella, and sports ball, among others.
Furthermore, in comparison with MCTformer, a method using Transformer, our approach attains competitive outcomes.
Harnessing the capabilities of the ResNet-38 backbone, our PLDA technique eclipses MCTformer by substantial margins, manifesting as mIoU values elevated by 19.4\%, 14.9\%, and 4.6\% for the frisbee, kite, and skis categories, respectively.
This resoundingly underscores its preeminence within these domains.
Remarkably, it is worth noting that both our PLDA method and the aforementioned counterparts encounter challenges in successfully segmenting the toaster and hair drier classes, highlighting the intricate nature of weakly supervised semantic segmentation tasks.

\subsection{Qualitative Analysis}

\subsubsection{Analysis of Feature Space}

\begin{figure}[t]
    \centering
    \includegraphics[width=0.48\textwidth]{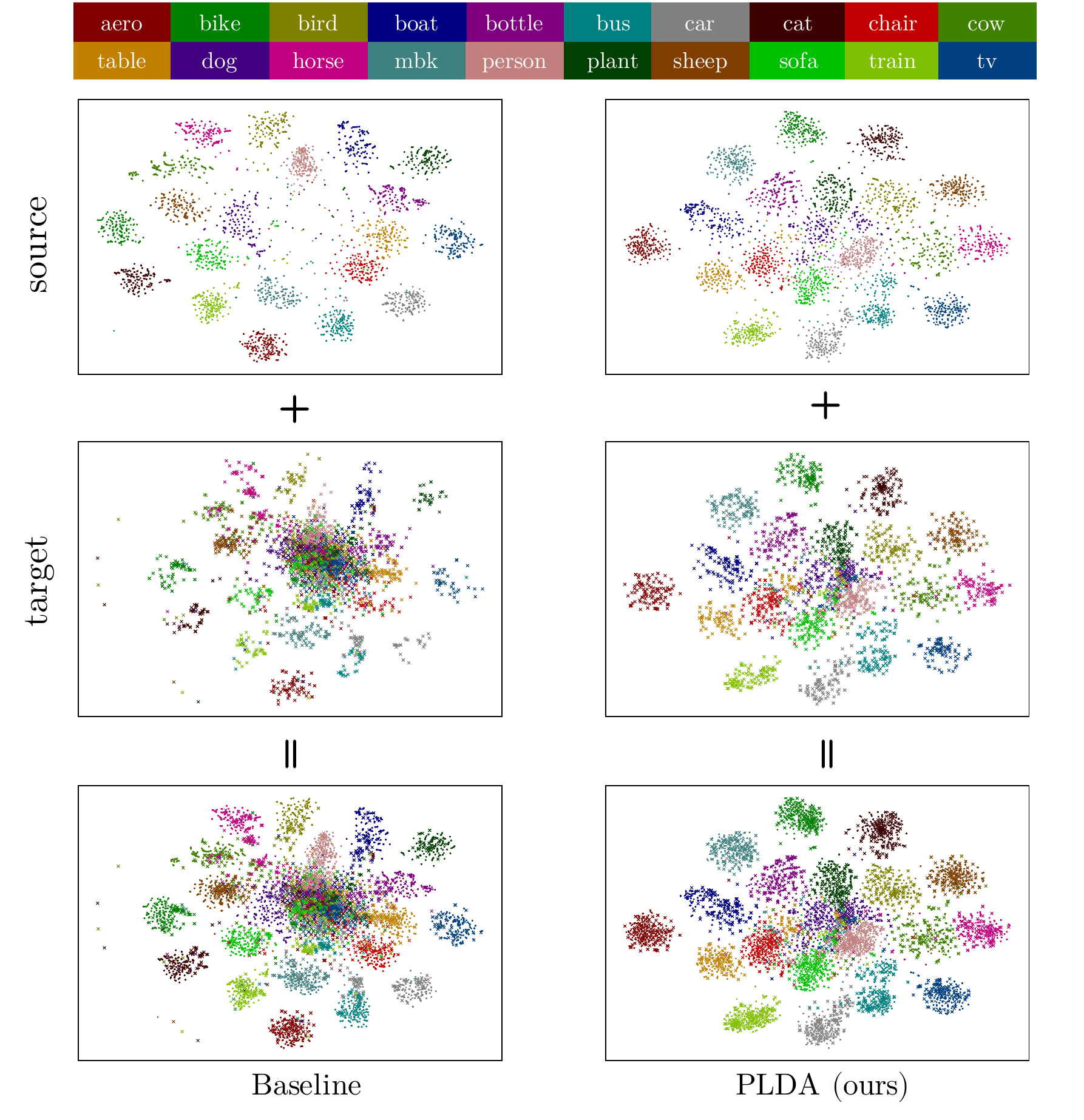}
    \vspace{-1em}
    \caption{
         \textbf{Qualitative Comparison} of t-SNE~\cite{van2008visualizing} visualization between the SEAM baseline and our PLDA method on PASCAL VOC 2012 dataset.
         ``." and ``×" represent source and target domain pixels, respectively.
         }
    \label{fig:tsne}
    \vspace{-10pt}
\end{figure}

Illustrated in \cref{fig:tsne} is the visualization of the feature space, employing the t-SNE methodology introduced in \cite{van2008visualizing}. Note that we first visualize the pixels in the source and target domains separately, as depicted in the first and second rows, respectively. The criteria for this division are established based on the activation value of the CAM. Specifically, pixels boasting an activation value surpassing 0.6 are deemed representative of the source domain.

It is clear that the intrinsic ability of the feature spaces associated with both the baseline method and our proposed PLDA to differentiate between categories within the source domain. However, a stark contrast emerges when inspecting pixel features emanating from the target domain under the purview of the baseline method: these features exhibit a propensity towards ambiguity, as they appear to be tightly confined within a localized spatial region. Such a result strongly alludes to an insufficiency in the acquired target domain features to proficiently address the segmentation task at hand. In direct contradistinction, our PLDA method demonstrates the class-wise discrimination ability in target domain features.

The third row in \cref{fig:tsne} shows the combination of the source and target domains. Compared to the baseline method, it is evident that the pixel features derived from our method exhibit a relatively close proximity in the feature space for both domains. This result serves to further corroborate the underpinning hypothesis of PLDA and its rationale.

\subsubsection{Semantic Segmentation Map Visualization}
\begin{figure*}[t]
    \centering
    \vspace{-0.5em}
    \includegraphics[width=0.99\textwidth]{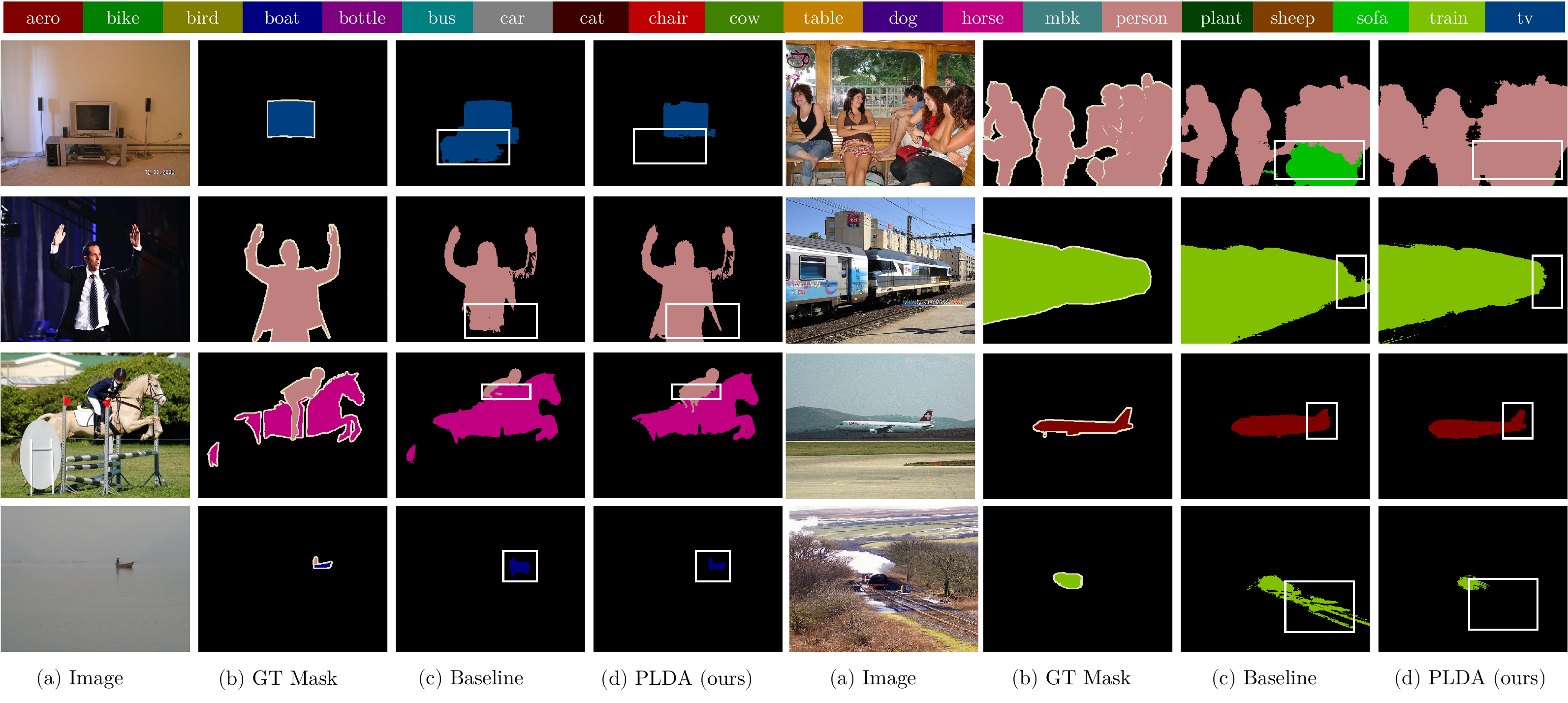}
    \vspace{-1em}
    \caption{\textbf{Qualitative comparisons} between the baseline and our PLDA method.
    The segmentation maps are obtained by DeepLabV1 with the ResNet38 backbone.
    The images are from the PASCAL VOC 2012 validation set.
    The white boxes highlight visual improvements compared to the baseline model.
    }
    \label{fig:seg1}
\end{figure*}

In this subsection, we present a visualization of the semantic segmentation maps generated by our PLDA method, juxtaposed with those generated by the baseline model, as depicted in \cref{fig:seg1}.
To get these visualization outcomes, we employ a ResNet38-based DeepLabV1 architecture \cite{wu2019wider}, which is trained employing pseudo labels generated by either the SEAM model or our innovative PLDA approach.
It is noteworthy that the delineated white bounding boxes serve to accentuate the visual improvements introduced by our method over the baseline model.
A discerning analysis of the figures unmistakably showcases the remarkable prowess of our PLDA method in rendering superior semantic segmentation maps within scenarios encompassing both intricate and elementary object configurations.
Significantly, our method demonstrates a notable capability in generating more comprehensive masks and effectively mitigating the incidence of erroneous predictions.
These visualization results eloquently establish the efficacy of our method in yielding enhanced semantic segmentation maps, thereby substantiating its inherent merits.

\section{Conclusion and {Discussion} }

In this paper, we revisit the imbalance activation issue in WSSS from an intra-image distribution discrepancy perspective and propose a simple yet effective Pixel-Level Domain Adaptation method to address it.
The key insight of our method is to diminish the distribution discrepancy by prompting the emergence of pixel features that are invariant with respect to the shift between the source and target domains.
At the same time, a Confident Pseudo-Supervision strategy is introduced to guarantee the semantic discriminability for pixel classification.
Extensive experiments demonstrate the effectiveness of our approach under a wide range of settings.

{
However, several directions merit further exploration. 
First, some components of PLDA, while considered pragmatic choices, may not be optimal. Particularly noteworthy is the challenge of accurately identifying source and target domain pixels within an image, a pressing concern given the absence of pixel-level annotations in WSSS. The CAM-based domain assignment strategy proposed herein employs a static threshold, determined via an ablation experiment, which may not be optimal across training iterations. 
Thereby, the adoption of a more adaptive domain assignment strategy could potentially enhance performance further, which we consider as interesting future work.
Second, our work provides a unified perspective to improve the integrity and quality of CAM. It possesses a generalizability that could be extended to other tasks, including \textit{CAM-based} object localization~\cite{Kim_2022_CVPR}, few-shot semantic segmentation~\cite{lee2022pixel}, and incremental semantic segmentation~\cite{Cermelli_2022_CVPR}. We believe that extending our approach to other tasks warrants future research endeavors, and we aspire for our methodology to serve as a catalyst for studies on new algorithms, analyses, and applications.}

\bibliographystyle{IEEEtran}
\bibliography{main}

\end{document}